\documentclass{aer} 
\usepackage{mathtools}

\usepackage[super,comma,round]{natbib}
\usepackage[figuresleft]{rotating}
\usepackage{txfonts}%
\usepackage{gensymb}
\usepackage{ragged2e}
\usepackage{tabularx,booktabs,caption}
\usepackage{subfigure}
\usepackage{textgreek}
\usepackage[ruled,vlined,linesnumbered]{algorithm2e}
\usepackage{graphicx}
\usepackage{color,xcolor,colortbl}
\usepackage{soul}
\usepackage{outlines,float}
\usepackage{blindtext}
\usepackage{comment}
\usepackage{hyperref,trees,geometry,amsmath,pgfplots}
\usepackage{adjustbox}
\usepackage{float}
\restylefloat{table}
\usepackage{lscape,tikz,tikzpeople,fancybox,wrapfig,epsfig,amssymb,multicol,enumerate,yhmath,pgfplots}
\usepackage{tkz-graph}
\usepackage{pgf-pie}
\usepackage{adjustbox}
\usepackage{wasysym}
\usetikzlibrary{shadows}
\usetikzlibrary{shapes.callouts}
\usetikzlibrary{matrix}
\pgfplotsset{compat=1.8}
\usepgfplotslibrary{statistics}
\usetikzlibrary{fadings}
\usetikzlibrary{arrows.meta}

\begin{document}

\markboth{da Silva Junior et al.}{A Systematic Study on Solving Aerospace Problems Using Metaheuristics}


\title{A Systematic Study on Solving Aerospace Problems Using Metaheuristics} 


\author{C. A. da Silva Junior \email{carlosdamat@ufsj.edu.br}}
\affil{Universidade Federal de S\~ao Jo\~ao del-Rei - UFSJ \\ Depertamento de Matem\'atica e Estat\'istica - DEMAT \\ S\~ao Jo\~ao del Rei - MG \\ Brazil}

\author{M. A. Pereira\email{marconit@ufsj.edu.br}}
\affil{Universidade Federal de S\~ao Jo\~ao del-Rei - UFSJ \\ Depertamento de Tecnologia em Engenharia Civ\'il, Computa\c{c}\~ao e Humanidades - DETEC \\ S\~ao Jo\~ao del Rei - MG \\ Brazil}

\author{A. Passaro\email{angelopassaro@gmail.com}\thanks{author3}}
\affil{Instituto de Estudos Avan\c{c}ados - IEAv \\ S\~ao Jos\'e dos Campos - SP \\ Brazil}

\author{C. A. da Silva Junior and A. Passaro}
\affil{Instituto Tecnol\'ogico de Aeron\'autica - ITA \\ PG-CTE \\ S\~ao Jos\'e dos Campos - SP \\ Brazil}

\maketitle                   

\begin{abstract}
Complex engineering problems can be modelled as optimisation problems. For instance, optimising engines, materials, components, structure, aerodynamics, navigation, control, logistics, and planning is essential in aerospace. Metaheuristics are applied to solve these optimisation problems. The present paper presents a systematic study on applying metaheuristics in aerospace based on the literature. Relevant scientific repositories were consulted, and a structured methodology was used to filter the papers. Articles published until March 2022 associating metaheuristics and aerospace applications were selected. The most used algorithms and the most relevant hybridizations were identified. This work also analyses the main types of problems addressed in the aerospace context and which classes of algorithms are most used in each problem.
\end{abstract}


\section*{NOMENCLATURE} 
\begin{deflist}
\listterm{ACR}{Aerospace Research Central}
\listterm{AIAA}{American Institute of Aeronautics and Astronautics}
\listterm{ACO}{Ant Colony Optimization}
\listterm{ABC}{Artificial Bee Collony}
\listterm{ANN}{Artificial Neural Network}
\listterm{BA}{Bat Algorithm}
\listterm{DE}{Differential Evolution}
\listterm{GTE}{Gas Turbine Engines}
\listterm{GA}{Genetic Algorithm}
\listterm{GWO}{Gray Wolf Optimizer}
\listterm{GLS}{Guided Local Search}
\listterm{HS}{Harmony Search}
\listterm{HM}{Heuristic Method}
\listterm{ILS}{Iterated Local Search}
\listterm{NSGA-II}{Non-dominated Sorting Genetic Algorithm II}
\listterm{MOGA}{Multi-objective Genetic Algorithm}
\listterm{PSO}{Particle Swarm Optimization}
\listterm{SA}{Simulated Annealing}
\listterm{SPEA2}{Strength Pareto Evolutionary Algorithm 2}
\listterm{TS}{Tabu Search}
\listterm{UAV}{Unmanned Aircraft System}
\listterm{VND}{Variable Neighborhood Descent}
\listterm{VNS}{Variable Neighborhood Search}
\listterm{WOA}{Whale Optimization Algorithm}
\end{deflist}

%
%

\section{INTRODUCTION}
\par Optimization is an intrinsic process to humanity. For instance, optimizing water extraction and finding a better feeding ground were essential for survival. Nowadays, optimizing manufacturing processes, design of devices and vehicles, properties of synthesized materials, energy distribution networks, and cellular antenna networks are pursued to guarantee, for instance, maximum profit, minimum energy consumption, and maximum coverage. In particular, computational optimization techniques have been extensively exploited to help solve several problems in Defence and Aerospace activities, such as in logistics, air transport of people and cargo, green aviation, border patrol, air combat, remotely piloted aircraft systems, and control of autonomous drone swarm. For instance, aircraft sequencing and scheduling problems in terminal airspaces \cite{2020CecenA} were solved using integer linear programming with meta-heuristics, and aircraft conflict resolution problems \cite{2020CecenB} using specialized heuristics.

\par As the problems to be solved become more complex, the number of design variables increases and global optimization approaches are required. Stochastic methods have been used frequently to obtain good and robust solutions in an acceptable time. Heuristics can be understood as rules (strategies) to solve problems. Regarding stochastic methods, metaheuristics stand out because of the cost/benefit relation. Metaheuristics are search methods defined as high-level methodologies (models), which can be used as guiding strategies in the design of underlying heuristics to solve specific optimization problems \cite{2009Talbi}. While metaheuristics get good solutions to exponential time problems in polynomial time, they are generally simple to implement and require relatively modest computational resources.

\par Many metaheuristics have been proposed in the literature, exploiting different ways to search for the problem extreme. Several metaheuristics present intrinsic parameters which control the search process. Such intrinsic parameters need a tuning procedure for solving adequately and efficiently the problem in focus. Such a multitude of methods makes it difficult to choose the best options to solve a given problem. Additionally, there are several applications in literature in which metaheuristics-based hybridization are exploited.

\par Metaheuristics are very popular, being made available implementations in libraries and frameworks in the most varied programming languages.
This paper presents a systematic review of optimization problems studied in the aerospace area using metaheuristics, given a glimpse of the main issues approached and how metaheuristics are used to solve distinct classes of problems. Moreover, it can guide the development of new algorithms and computational tools. We emphasize that we do not focus on the type of modelling or even on the way in which each research group gets involved with the most diverse types of problems in the aerospace area. The focus of this work is to evaluate how meta-heuristics are being used to solve problems in the area.

\par The main contributions of this paper are:
\begin{itemize}
	\item a systematic review of the current studies of last decade that approaches aerospace problems using metaheuristics;
	
	\item the identification of the main metaheuristics used in the area;
	
	\item the identification of the main problems where the metaheuristics are applied;
	
	\item the mapping of the application of the main metaheuristics to problems in the field of aerospace engineering.
	
\end{itemize}

\par This paper is structured as follows: In Section \ref{sec:Background} the related works are presented. In Section \ref{sec:Methodology}, the methodology used to collect, classify, and analyse the papers is detailed. Section \ref{sec:ResultsAndDiscussion} discusses the selected works, classifying the articles and their objectives, details the metaheuristics addressed and the problems to which these algorithms were applied. The conclusions are presented in Section \ref{sec:Conclusao}.

\section{BACKGROUND}\label{sec:Background}

\par The application of metaheuristic optimization algorithms in engineering problems is very po\-pu\-lar. Many reviews have been produced to identify how the various research groups have used metaheuristics in their respective areas of interest.

\par Gao et al. \cite{2020Gao} present a study on applying metaheuristics for solving disassembly line balancing, planning and scheduling problems in remanufacturing. Molina et al. \cite{2020Molina} present a review on metaheuristic optimization in controller tuning problem. Similarly, Tayarani-N. et al. \cite{2015Tayarani} provide a study on Metaheuristic Algorithms in Car Engine Design. In these three studies mentioned, the authors look for sub-areas within the context of the chosen problems, as well as to identify which are the metaheuristics best applied to each one of these sub-problems.

\par Huang et al. \cite{2020Huang} focus on parameter tuning methods for metaheuristics. Despite their generality and applicability in a huge diversity of problems, each application area usually demand a different parameter configuration. The authors present three categories of parameter tuning and discuss the strengths and weaknesses of each one of them. Rodrigues et al. \cite{2017Rodrigues} present a study on mono and multi-objective metaheuristics to solve machine learning problems. In this case, a machine learning problem can be viewed as an optimization problem, mono or multi-objective.

\par The small list of articles briefly depicted illustrates the amplitude of applications and studies accomplished with metaheuristics. Metaheuristics have also been applied to solve problems in aerospace.

\par Montano et al. \cite{2012Montano} present a review on the application of multi-objective optimization algorithms in aeronautical and aerospace, particularly addressing the conceptual and preliminary design phases. Methods like Multi-objective Genetic Algorithm (MOGA), Strength Pareto Evolutionary Algorithm 2 (SPEA2), Non-dominated Sorting Genetic Algorithm II (NSGA-II), Particle Swarm Optimization (PSO), and Differential Evolution (DE) are briefly described, presenting their applications. The authors classify the use of the methods in the following classes: Conceptual Design Optimization, 2-D Geometries and Airfoil Shape Optimization, 3-D Complex Physics/Shape Optimization, Structural Optimization, Multidisciplinary Design Optimization, Aerospace System Optimization, Control System Design. Such classification takes into account the complexity degree of the optimization problem. The authors emphasize some relevant characteristics of multi-objective evolutionary algorithms, such as robustness, multiple solutions per run, ease of parallelizing and hybridizing, and simplicity. The high computational cost is indicated by several researchers as a relevant disadvantage.

\par Jafari and Nikolaidis \cite{114} focus their study on using metaheuristic global optimization for aircraft engine modelling and controller design. They present a systematic examination of the critical areas in gas turbine engines, GTE, modelling, control structure design, and controller tuning to highlight the remaining research challenges for metaheuristic algorithms in new advanced applications. Two approaches for global optimization algorithms are proposed to deal with future challenges in the area: metaheuristics combined with advanced computational methods, such as learning techniques, and hybrid optimization algorithms combining metaheuristics and gradient-based algorithms.

\par Wu \cite{032} present a survey focused on population-based metaheuristics for motion planning of aircraft. He assesses classic algorithms as well as their variations applied to the context. Aspects such as population size, dynamics of movement/updating of the population, strategies for evaluating individuals are discussed, as well as the application of each category of metaheuristics to the problem under study.

\par Bashab et al. \cite{2020Bashab} present a study on the application of metaheuristics to solve university timetabling problems. The work was conducted using the systematic literature review detailed in \cite{2007Keele}. The authors carried out a systematic study of the literature, identifying and classifying the different works that address a solution of the problem in focus using metaheuristics. The study classifies the papers into categories such as: (a) New method: presents a new algorithm; (b) Method: uses some algorithm already consolidated in the literature; (c) Modified method: it presents an algorithm with some specific modification; (d) Hybrid method: integrates two or more algorithms; (e) Review: performs the examination of an algorithm within an evaluation process; (f) Survey: performs an evaluation of an algorithm to produce a description of the main characteristics of the studied metaheuristics; (g) Case Study: performs the evaluation of one or more algorithms using a real problem; among others.

\par We propose in this paper a study focused on the application of metaheuristics in aerospace area. Our approach is similar to that adopted by Bashad et al. \cite{2020Bashab}.

\section{METHODOLOGY}\label{sec:Methodology}

\par Our research is based on data from scientific repositories: SCOPUS, IEEE and Web of Science. These bases include articles from journals, congresses, and from relevant organizations in the aerospace context, such as the American Institute of Aeronautics and Astronautics (AIAA) Aerospace Research Central (ARC).

\par The first searches were carried out considering the following parameters:
\begin{itemize}
	\item Publication date between 2000 and march of 2022;
	\item Combination of keywords of two group:
	\begin{itemize}
		\item Group 1: “aerospace”, “aeronautical”, “aircraft”, “UAV”, “unmanned aerial vehicle”;
		\item Group 2: “metaheuristic”, “meta-heuristic”;
	\end{itemize}
	\item Were excluded results associated with books, book chapters, patents, notes and editorials.
\end{itemize}

\par The words of Group 1 were chosen to obtain an overview of the work developed in the aerospace area, without privileging a specific line of research. Some additional terms were also considered in the initial stages of this research. For example, including the term “drone” in the SCOPUS database, there was an increase of 50 articles, i.e, an increase of 14.71\% in the articles collected. However, 39 of these papers were published after 2018 and focus on logistical issues. One of these papers, Deng et. al \cite{2020Deng}, optimize a vehicle routing problem with motion synchronization of drones, side-walk robots or pedestrians. This type of problem is encompassed in UAV studies. We chose to limit the search to a small number of relevant keywords in the aerospace context.

\par The keywords of Group 2 were chosen to evaluate how metaheuristics have been used in the areas presented in Group 1. It is worth highlighting that the optimization field is very broad, and other lines of analysis can be chosen, such as direct methods and heuristics. To illustrate, using the term ``optimization'' in keywords of Group 2, instead of metaheuristics and their variations, in the same analysis period presented here would generate 100,830 articles in phase 1.

\par The logical expressions used to proceed the searches in the databases are detailed in Table \ref{tab:AdvancedSearchDatabase}. The initial search is processed using the title, the abstract and the keywords of the papers. After the initial search, a set of filters is used in four phases. In the first phase, all documents related to scientific meetings that do not correspond to a full paper were excluded. In the second phase, duplicated papers, i.e., articles found in multiple databases, were excluded. In the third phase, we excluded articles that were not published in English and the ones we had no access to the full paper. In the last phase, papers that did not explicitly present the keywords registered in Table \ref{tab:AdvancedSearchDatabase} in the article's body were removed.
\begin{table}[H]
	\centering
	\caption{Logical expression used in the advanced search module in each database.}
	\begin{tabular}{c}
		\hline
		\href{https://www.scopus.com/}{SCOPUS} \\\hline
		(TITLE\_ABS\_KEY(aerospace) OR TITLE\_ABS\_KEY(aeronautical) OR TITLE\_ABS\_KEY(aircraft) \\
		OR TITLE\_ABS\_KEY(UAV) OR TITLE\_ABS\_KEY(unmanned AND aerial AND vehicle)) \\
		AND \\ 
		(TITLE\_ABS\_KEY(metaheuristic) OR TITLE\_ABS\_KEY(meta-heuristic))  \\ \hline
		\href{https://ieeexplore.ieee.org/}{IEEE} \\ \hline
		(``All Metadata'':``aerospace'' OR ``All Metadata'':``aeronautical'' OR ``All Metadata'':``aircraft'' OR \\
		``All Metadata'':``UAV'' OR ``All Metadata'':``unmanned aerial vehicle'') \\
		AND \\
		( ``All Metadata'':``metaheuristic'' OR ``All Metadata'':``meta\--heuristic'' ) \\ \hline
		\href{https://www.webofknowledge.com/}{Web Of Science} \\ \hline
		(All = (aerospace OR aeronautical OR aircraft OR \\
		UAV OR (unmanned AND aerial AND vehicle)) ) \\
		AND \\
		( All = (metaheuristic OR meta\--heuristic ) ) \\ \hline
	\end{tabular}
	\label{tab:AdvancedSearchDatabase}
\end{table}

\par In the next section, the process and data collected are analysed in order to highlight, for instance, how and how many metaheuristics have been used, the most used ones, the kind of problems in aerospace that have been focused on, in addition to other relevant aspects.

\section{RESULTS AND DISCUSSIONS}\label{sec:ResultsAndDiscussion}

\par The selected papers cover many applications and different approaches for solving optimization problems. Before initiating the analysis, we present a short resume of the focus of five filtered articles. In the first filtering phase, a search was carried out using the chosen keywords presented in Table 1, highlighting “meta-heuristics”, which is the study’s central point, followed by the context in which “aerospace” is being applied. In the second phase, books and conference proceedings were eliminated to focus on journal articles. In the third filtering phase, duplicate articles were eliminated so that each article was not counted more than once in the search. In the fourth filtering phase, articles that were not in English or for which access was denied were eliminated. Finally, we selected journals in which the authors highlighted meta-heuristics as a central point in their research, as this is the main object of search and analysis to be verified in this research. This  made it possible to obtain a reasonable number of journals to analyse.

\par A comparative study of meta-heuristic algorithms for solving UAV path planning is presented in \cite{003}. The authors model the problem as a constrained single objective optimization problem and compare the performance of seven well-known standard metaheuristic algorithms. The goal is to study the behaviour and efficiency of the optimization algorithms under different conditions for the UAV path planning problem. The authors state that the approaches described in the paper can be generalized to other classes of trajectory planning problems, such as the problem of planning UAV trajectories in three-dimensional space.

\par Mansi et al. \cite{069} address the disruptions in the airline industry by different causes (e.g., cancelled flights, modifications of the number of possible landings/take-off at some airports), recovery strategies (e.g., cancelling flights, splitting groups of passengers into sub-groups and rescheduling these sub-groups on different alternative flights) and many practical constraints related to the aircraft (e.g., maximum distances, maintenance), the passengers (e.g., maximum acceptable delay) and the airports (e.g., capacity). The authors propose a hybrid method that uses metaheuristics with an oscillation strategy and mixed integer programming. The decision variables are mostly binary but include at least one continuous variable.

\par The pickup and delivery problem with timing constraints is dealt with in \cite{146}. The article focuses on the availability of regular, quick, and cost-effective transport in Bangalore, but the problem strongly relates to the aerospace context. The authors use three population-based algorithms and one single-solution-based metaheuristic, and the problem is formulated using Boolean and continuous decision variables. This paper survived the filtering phases because one of the authors is affiliated with an aerospace institution. 

\par Sarangkum et al. \cite{052} apply multi-objective evolutionary optimization to design aircraft fuselage stiffeners. The optimization model searches to minimize the structural mass and compliance and maximize the first-mode natural frequency of the structure. The decision variables are continuous. The authors compare the results of three multi-objective metaheuristics.

\par Sopov \cite{030} proposes a self-configuring genetic algorithm for multimodal optimization problems. The binary representation of decision variables is adopted, and the authors claim that the approach can be implemented for many real-world problems with mixed types. The algorithm is evaluated by using benchmark problems. Once again, this article survived the filtering phases because the author is with an aerospace institution.

\par In the next sections we highlight some aspects regarding the use of metaheuristics to solve aerospace problems. The basis is the content of the articles selected after the four filtering phases. The analysis makes use of the terminology adopted in each selected article. Therefore, the names, classifications, type of optimization problem, decision variables follow the one presented in the articles.

\subsection{Selection of the papers}\label{sec:Search}
\par The search procedure results in 1092 pre-selected papers. Notice that each database carries out the advanced search using its own algorithm. These articles were published in 611 different  Journals/Proceedings: 55.7\% are journal articles, 36.9\% are Conference papers, 4.2\% are articles in Early Access, and 3.2\% are Reviews. The number of articles per year in each database is presented in the Table \ref{tab:QuantityOfPublicationsForYears}. The searches were carried out in March/2022, so the data corresponding to 2022 only include articles available in the first two months of this year.
\begin{table}[H]
	\centering
	\caption{Number of publications in each database.}	\label{tab:QuantityOfPublicationsForYears}
	\begin{adjustbox}{width=\columnwidth,center}
		\begin{tabular}{ccccccccccccccccccccccc} \hline \\ \\
			\begin{rotate}{90}\textbf{Year}\end{rotate} & \begin{rotate}{90}\textbf{2002}\end{rotate} & \begin{rotate}{90}\textbf{2003}\end{rotate} & \begin{rotate}{90}\textbf{2004}\end{rotate} & \begin{rotate}{90}\textbf{2005}\end{rotate} & \begin{rotate}{90}\textbf{2006}\end{rotate} & \begin{rotate}{90}\textbf{2007}\end{rotate} & \begin{rotate}{90}\textbf{2008}\end{rotate} & \begin{rotate}{90}\textbf{2009}\end{rotate} & \begin{rotate}{90}\textbf{2010}\end{rotate} & \begin{rotate}{90}\textbf{2011}\end{rotate} & \begin{rotate}{90}\textbf{2012}\end{rotate} & \begin{rotate}{90}\textbf{2013}\end{rotate} & \begin{rotate}{90}\textbf{2014}\end{rotate} & \begin{rotate}{90}\textbf{2015}\end{rotate} & \begin{rotate}{90}\textbf{2016}\end{rotate} & \begin{rotate}{90}\textbf{2017}\end{rotate} & \begin{rotate}{90}\textbf{2018}\end{rotate} & \begin{rotate}{90}\textbf{2019}\end{rotate} & \begin{rotate}{90}\textbf{2020}\end{rotate} & \begin{rotate}{90}\textbf{2021}\end{rotate} & \begin{rotate}{90}\textbf{2022}\end{rotate} & \begin{rotate}{90}\textbf{Total}\end{rotate} \\ \hline
			\textbf{Scopus}         & 1 & 1 & 4 & 3 & 6 & 4 & 14 & 10 & 7 & 13 & 21 & 11 & 16 & 14 & 20 & 37 & 48 & 60 & 63 & 93 & 11 & \textbf{457} \\
			\textbf{IEEE}           & 0 & 2 & 1 & 2 & 0 & 2 &  1 &  2 & 1 & 7  &  4 &  7 & 11 &  8 &  6 & 16 & 16 & 19 & 27 & 35 &  2 & \textbf{169} \\
			\textbf{Web of Science} & 1 & 1 & 1 & 1 & 5 & 8 &  8 &  4 & 6 & 12 & 17 & 16 & 15 & 26 & 38 & 47 & 47 & 65 & 56 & 67 & 25 & \textbf{466} \\ \hline
		\end{tabular}
	\end{adjustbox}
\end{table}

\par
\begin{figure}[H]
	\centering
	\includegraphics[width=0.75\textwidth]{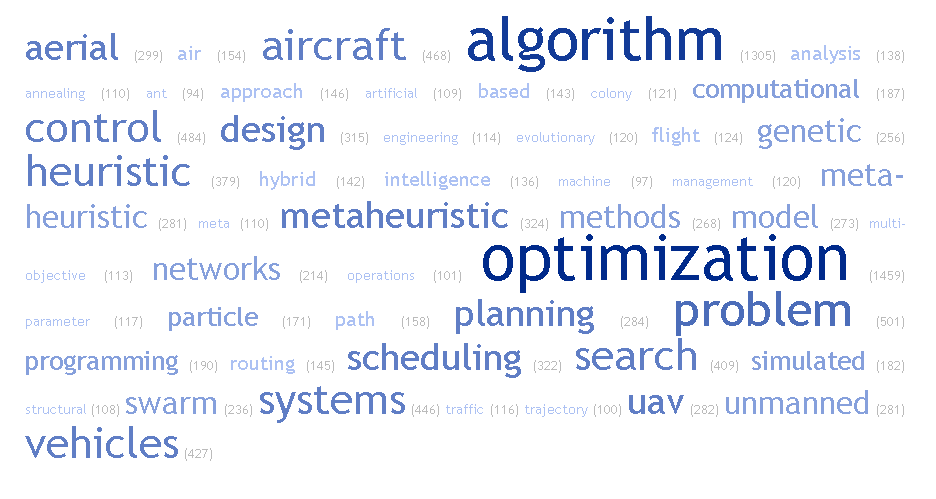}
	\caption{Word cloud built using the first 50 most frequent words in the title and keywords of the article defined by the authors and the journal.}
	\label{fig:WordCloud}
\end{figure}

 Figure \ref{fig:WordCloud} presents a word cloud produced by the junction of the results obtained from SCOPUS, IEEE and Web of Science after removing repeated articles (step 2). Only the text of the title and the list of keywords were used. We considered the keywords provided by the authors and the keywords provided by the journals/publishers. As expected, the word optimization shows prominence, followed by the word algorithm, problem, and control heuristic. Metaheuristics could be classified in a second group, together with some applications, such as planning, scheduling, aircraft, systems and vehicles. Some metaheuristics appear in a third level: genetic, swarm, simulated. That suggests that, generally, the titles and keywords of the articles in the aerospace area are mainly concerned with the problem to be solved and not necessarily with the technique used.

\par The article selection process and the remaining number of documents at the end of each filtering phase are presented in Figure \ref{fig:QuantityOfPublications}. In the 1st phase, 105 conference books and proceedings were eliminated, leaving 987 works. In Phase 2, 58 duplications were eliminated, remaining 929 papers. In the 3rd phase, 34 articles were removed, leaving 895 articles. In the 4th phase, 697 papers were eliminated, leaving a manageable number of 198 articles.
\begin{figure}[H]
	\centering
	\begin{tikzpicture}[scale=0.3]
		\draw[fill=green!50, thick] (0, 0) circle (5);
		\draw (-7.5,4.8) -- (0.0,4.8);
		\draw node at (-7.5,5.3) {Search};
		\draw node at (-6.9,4.3) {1092};
		\draw[fill=red!30, thick] (0, 0) circle (4.52);
		\draw (7.5,3.8) -- (0.0,4.48);
		\draw node at (7.5,4.3) {1st Phase};
		\draw node at (6.9,3.3) {987};
		\draw[fill=blue!30, thick] (0, 0) circle (4.25);
		\draw (-7.5,2.6) -- (0.0,4.15);
		\draw node at (-7.5,3.1) {2nd Phase};
		\draw node at (-6.9,2.0) {929};
		\draw[fill=orange!30, thick] (0, 0) circle (4.01);
		\draw (7.5,1.9) -- (0.0,1.7);
		\draw node at (7.5,2.4) {3rd Phase};
		\draw node at (6.9,1.3) {895};
		\draw[fill=black!30, thick] (0, 0) circle (0.9);
		\draw (-7.5,0.0) -- (0.0,0.0);
		\draw node at (-7.5,0.6) {4th Phase};
		\draw node at (-6.9,-0.5) {198};
	\end{tikzpicture}
	\caption {Articles remaining after each filtering phase.}
	\label{fig:QuantityOfPublications}
\end{figure}

\par
\begin{figure}[H]
	\centering
	\begin{tikzpicture}[scale=0.5]
		\begin{axis}[
			ybar,
			bar width=3pt,
			height=10cm,
			width=25cm,
			axis on top,
			y axis line style={opacity=0},
			axis x line*=bottom,
			tick align=inside,
			tick style={draw=none},
			xtick=data,
			ymajorgrids,
			major grid style={draw=white},
			xtick={1,2,3,4,5,6,7,8,9,10,11,12,13,14,15,16,17,18,19,20,21},
			xticklabels={2002, 2003, 2004, 2005, 2006, 2007, 2008, 2009, 2010, 2011, 2012, 2013, 2014, 2015, 2016, 2017, 2018, 2019,2020,2021,2022},
			x tick label style={rotate=90,anchor=east},
			legend style={
				at={(0.0,-0.15)},
				anchor=north west,
				legend columns=-1,
				/tikz/every even column/.append style={column sep=0.5cm}
			},
			]       
			\addlegendimage{empty legend}
			\addlegendentry{\textbf{Legend:}}                   
			\addplot coordinates {(1, 1) (2, 1) (3, 4) (4, 3) (5, 6) (6, 4) (7, 14) (8, 10) (9, 7) (10, 13) (11, 21) (12, 11) (13, 16) (14, 14) (15, 20) (16, 37)
				(17, 48) (18, 60) (19, 63) (20, 93) (21, 11)}; \addlegendentry[text width=40pt, text depth=]{SCOPUS}
			\addplot coordinates {(1, 0) (2, 2) (3, 1) (4, 2) (5, 0) (6, 2) (7, 1) (8, 2) (9, 1) (10, 7) (11, 4) (12, 7) (13, 11) (14, 8) (15, 6) (16, 16)
				(17, 16) (18, 19) (19, 31) (20, 35) (21, 2)}; \addlegendentry[text width=40pt, text depth=]{IEEE}
			\addplot coordinates {(1, 1) (2, 1) (3, 1) (4, 1) (5, 5) (6, 8) (7, 8) (8, 4) (9, 6) (10, 12) (11, 17) (12, 16) (13, 15) (14, 26) (15, 38) (16, 47)
				(17, 47) (18, 62) (19, 59) (20, 67) (21, 25)}; \addlegendentry[text width=80pt, text depth=]{Web Of Science}
			
			\addplot[ultra thick,fill=black] 
			coordinates {(1, 0) (2, 0) (3, 1) (4, 0) (5, 2) (6, 4) (7, 4) (8, 6) (9, 4) (10, 5) (11, 7) (12, 4) (13, 5) (14, 16) (15, 13) (16, 18)
				(17, 22) (18, 23) (19, 27) (20, 32) (21, 7)}; \addlegendentry[text width=100pt, text depth=]{Papers after filtering}

		\end{axis} 
	\end{tikzpicture}
	\caption{Evolution of the number of articles found in each database by year. The black bar represents the evolution of the selected papers after filtering.}
	\label{fig:QuantityOfPublicationsForYears}
\end{figure}
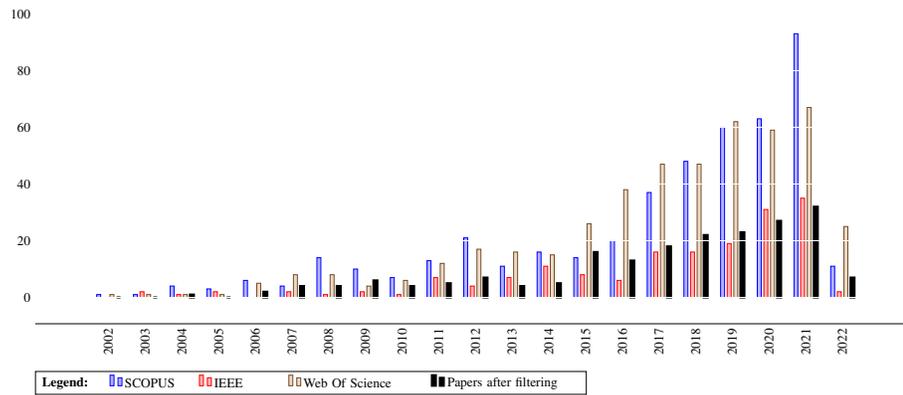

 The number of papers obtained by the search process by year from each consulted repository, Table \ref{tab:QuantityOfPublicationsForYears}, is shown in Figure \ref{fig:QuantityOfPublicationsForYears}, as well as the number of papers that remains after the filtering phases. The profile is similar, evidencing the growth of metaheuristics used in aerospace.

\subsection{Papers Contributions}\label{sec:Papers}

\par In the first analysis, the articles were classified into three groups based on the declared purpose of the paper:
\begin{enumerate}
	\item Performance analysis: the main objective is to evaluate the optimization algorithms, encompassing around 55\% of the selected papers;
	\item Case studies: the main objective is to solve a specific problem using one or more optimization algorithms; about 36,9\% of the selected papers fall into this group;
	\item Revision and Surveys: the objective is to present a critical analysis or a general view of the tendencies to perform the study on a particular topic of interest; they sum about 8.1\% of filtered papers.
\end{enumerate}

\par To illustrate the content of the first group, we cite Yang et al. \cite{001}, which analyses the performance of sixteen metaheuristics for solving a mixed integer commitment problem. Notice that the problem in which the metaheuristics were applied is just a pretext for the analysis. For the second group exemplification, Hbaieb et al. \cite{005} focus on optimizing the scheduling of aircraft landings at a given airport which is the article's main focus. Table \ref{tab:TableTiposDePapers} registers the selected references associated with each class.
\begin{table}[H]
	\centering
	\caption{Distribution of papers by class} 
	\label{tab:TableTiposDePapers}       
	\begin{center}
		\begin{adjustbox}{width=\textwidth}
			\begin{tabular}{l|l}
				\hline
				\textbf{Class}        & Papers  \\ \hline\hline
				& \cite{146}, \cite{030}, \cite{001}, \cite{060}, \cite{027}, \cite{168}, \cite{110}, \cite{099}, \cite{203}, \cite{190},
				\cite{154}, \cite{209}, \cite{131}, \cite{045}, \cite{073}, \cite{070}, \cite{188}, \cite{138}, \\
				& \cite{057}, \cite{107}, \cite{088}, \cite{074}, \cite{091}, \cite{170}, \cite{182}, \cite{016}, \cite{021}, \cite{196}, 
				\cite{023}, \cite{161}, \cite{042}, \cite{207}, \cite{130}, \cite{134}, \cite{135},\\
				& \cite{211}, \cite{086}, \cite{162}, \cite{126}, \cite{055}, \cite{212}, \cite{083}, \cite{059}, \cite{159}, \cite{184}, 
				\cite{187}, \cite{094}, \cite{017}, \cite{125}, \cite{046}, \cite{013}, \cite{044}, \\
				Performance & \cite{165}, \cite{096}, \cite{143}, \cite{148}, \cite{186}, \cite{155}, \cite{145}, \cite{080}, \cite{129}, \cite{178}, 
				\cite{103}, \cite{218}, \cite{098}, \cite{009}, \cite{025}, \cite{171}, \cite{200}, \\
				Analysis    & \cite{156}, \cite{157}, \cite{193}, \cite{022}, \cite{117}, \cite{068}, \cite{214}, \cite{142}, \cite{014}, \cite{169}, 
				\cite{102}, \cite{201}, \cite{144}, \cite{195}, \cite{076}, \cite{121}, \cite{123}, \\
				& \cite{122}, \cite{160}, \cite{118}, \cite{216}, \cite{150}, \cite{191}, \cite{147}, \cite{004}, \cite{061}, \cite{012}, 
				\cite{194}, \cite{198}, \cite{141}, \cite{043}, \cite{205} \\
				& \cite{062}, \cite{174}, \cite{115}, \cite{175}, \cite{028}, \cite{116}, \cite{087}, \cite{089} \\ \hline\hline
				& \cite{003}, \cite{069}, \cite{052}, \cite{041}, \cite{106}, \cite{051}, \cite{109}, \cite{040}, \cite{075}, \cite{085}, 
				\cite{006}, \cite{119}, \cite{101}, \cite{108},\\
				& \cite{149}, \cite{019}, \cite{064}, \cite{063}, \cite{206}, \cite{066}, \cite{192}, \cite{215}, \cite{183}, \cite{050}, 
				\cite{136}, \cite{097}, \cite{010}, \cite{132}, \\
				Case Studies& \cite{039}, \cite{164}, \cite{100}, \cite{035}, \cite{007}, \cite{204}, \cite{219}, \cite{079}, \cite{139}, \cite{008}, 
				\cite{034}, \cite{185}, \cite{026}, \cite{113}, \\
				& \cite{011}, \cite{128}, \cite{015}, \cite{124}, \cite{067}, \cite{210}, \cite{078}, \cite{077}, \cite{179}, \cite{082}, 
				\cite{065}, \cite{024}, \cite{005}, \cite{047}, \\
				& \cite{033}, \cite{120}, \cite{153}, \cite{189}, \cite{056}, \cite{181}, \cite{166}, \cite{049}, \cite{084}, \cite{111}, 
				\cite{213}, \cite{197}, \cite{054}, \cite{038}, \\
				& \cite{072}, \cite{053}, \cite{020}, \cite{140} \\ \hline\hline
				Reviews or  & \cite{114}, \cite{032}, \cite{090}, \cite{048}, \cite{029}, \cite{031}, \cite{217}, \cite{173}, \cite{152}, \cite{036}, 
				\cite{137}, \cite{127}, \cite{172}, \cite{002}, \\
				Surveys     & \cite{058} \\ \hline\hline
			\end{tabular}
	\end{adjustbox}	\end{center}
\end{table}

\par
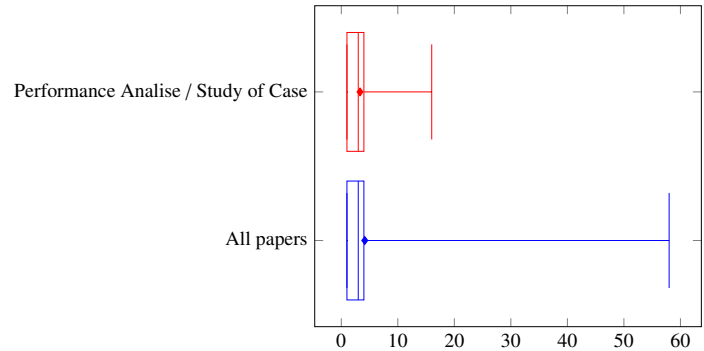
\begin{figure}[H]
	\centering
	\begin{tikzpicture}[scale=0.75]
		\begin{axis}
			[
			ytick={1,2},
			yticklabels={All papers, Performance Analise / Study of Case},
			]
			\addplot+[
			boxplot prepared={
				upper whisker=58.0,
				upper quartile=4.0,
				average=4.1566,
				median=3.0,
				lower quartile=1.0,
				lower whisker=1.0
			},
			] coordinates {};
			\addplot+[
			boxplot prepared={
				upper whisker=16.0,
				upper quartile=4.0,
				average=3.3157,
				median=3.0,
				lower quartile=1.0,
				lower whisker=1.0
			},
			] coordinates {};
		\end{axis}
	\end{tikzpicture}
	\caption{Box plot of the metaheuristics used in the selected papers. The review/survey articles show much more citations, but the median and average of the metaheuristics used are practically the same.}\label{fig:MediasAgoritmosPapers}
\end{figure}

 The metaheuristics used in the Performance Analysis and Case Studies groups range from one to eighteen, with an average of three metaheuristics per article. Considering all filtered articles, which include the Review/Research class, the average number of algorithms referenced per article increases to four, Figure \ref{fig:MediasAgoritmosPapers}, but the median of the metaheuristics cited is practically the same.

\par
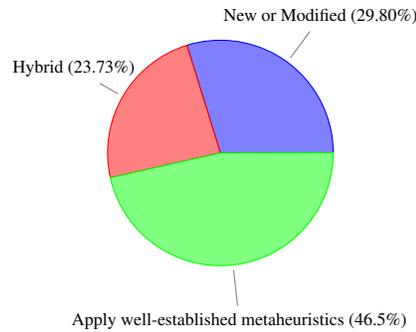
\begin{figure}[H]
	\centering
	\def\angle{0}
	\def\radius{2}
	\def\labelradius{3}
	\def\cyclelist{{"blue","red","green","gray","magenta","cyan","orange"}}
	
	\newcount\cyclecount \cyclecount=-1
	\newcount\ind \ind=-1
	\begin{tikzpicture}[nodes = {font=\small},scale=0.75,transform shape]
		\foreach \percent/\name in {
			29.80/New or Modified, 
			23.73/Hybrid, 
			46.5/Apply well-established metaheuristics, 
		} {
			\ifx\percent\empty\else               
			\global\advance\cyclecount by 1     
			\global\advance\ind by 1            
			\ifnum6<\cyclecount                 
			\global\cyclecount=0              
			\global\ind=0                     
			\fi
			\pgfmathparse{\cyclelist[\the\ind]} 
			\edef\color{\pgfmathresult}         
			\draw[fill={\color!50},draw={\color}] (0,0) -- (\angle:\radius)
			arc (\angle:\angle+\percent*3.6:\radius) -- cycle;
			\draw[draw=gray, shorten >=2pt] (\angle+0.5*\percent*3.6:\labelradius) node {\name~(\percent\%)} edge (\angle+0.5*\percent*3.6:\radius);
			\pgfmathparse{\angle+\percent*3.6}  
			\xdef\angle{\pgfmathresult}         
			\fi
		};
	\end{tikzpicture}
	\caption{Metaheuristics in the selected papers.}\label{fig:UsoDeMEtaheuristicas}
\end{figure}

 The second analysis focuses on the role of metaheuristics in the papers. It was identified that 29.8\% of the filtered articles are dedicated to exploring new or modified metaheuristics, and 23.7\% study hybrid metaheuristics. The remaining papers (46.5\%) solve problems applying well-established metaheuristics of the literature, such as Genetic Algorithms and Particle Swarm algorithms. Figure \ref{fig:UsoDeMEtaheuristicas} illustrates the complete image.

\par We identified 377 different metaheuristics in the selected papers. The metaheuristics were grouped into eighteen classes based on the inspiring metaphor, following the classification presented in the selected articles. The proposed families presented in Figure \ref{fig:ClassesDeAlgoritmos} and Table \ref{tab:TableFrequenciaClassAndMainMetaheuristcs} are Ant, Bat, Bee, DE (differential evolution), Evolutionary (based on genetic algorithms - GA), GLS (guided local search), heuristics, HS (Harmony Search), ILS (iterated local search), Learning, Neural, Simulated, PSO (based on Particle Swarm Optimization), Tabu (based on Tabu Search - TS), VND (Variable Neighbourhood Descent), VNS (Variable Neighbourhood Search), Whale and Wolf. The frequency of occurrence of each class in the filtered articles is shown in the second column in Table \ref{tab:TableFrequenciaClassAndMainMetaheuristcs}. Columns 3 and 4 identifies the main metaheuristic referenced in each class and the number of times this particular metaheuristic is used.
\begin{figure}[H]
	\centering
	\begin{tikzpicture}[scale=0.75]
		\begin{axis}[
			ybar,
			height=10cm,
			width=10cm,
			axis on top,
			y axis line style={opacity=0},
			axis x line*=bottom,
			tick align=inside,
			tick style={draw=none},
			xtick=data,
			ymajorgrids,
			major grid style={draw=white},
			xtick={1,2,3,4,5,6,7,8,9,10,11,12,13,14,15,16,17,18},
			xticklabels={Ant, Bat, Bee, DE, Evolutionary, GLS, Heuristics, HS, ILS, Learning, Neural, PSO, Simulated, Tabu, VND, VNS, Whale, Wolf},
			x tick label style={rotate=90,anchor=east},
			legend style={
				at={(0.0,-0.15)},
				anchor=east,
				legend columns=-1,
				/tikz/every even column/.append style={column sep=0.15cm}
			},
			]       
			\addlegendimage{empty legend}
			\addplot coordinates {(1, 47) (2, 19) (3, 24) (4, 45) (5, 217) (6, 3) (7, 110) (8, 25) (9, 28) (10, 29)	(11, 14) (12, 242) (13, 58) (14, 33) (15, 8) (16, 18) (17, 11) (18, 46)};
		\end{axis} 
	\end{tikzpicture}
	\caption{Frequency of each class of metaheuristics presented in the papers analysed.}\label{fig:ClassesDeAlgoritmos}
\end{figure}
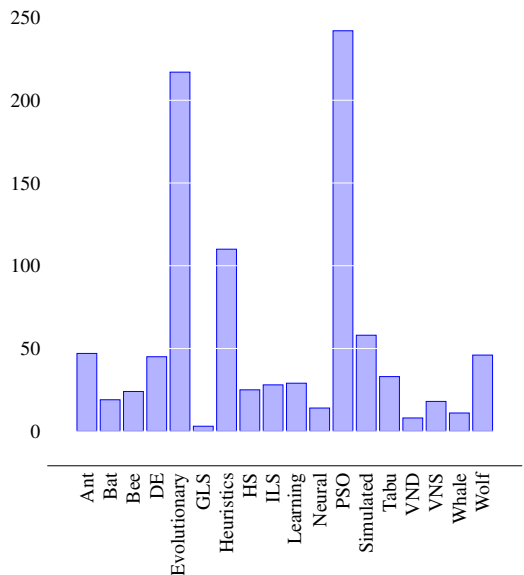
\begin{table}[H]
	\centering
	\begin{tabular}{cccc}
		\hline
		\multicolumn{2}{c|}{Family class}                                   & \multicolumn{2}{c}{Main metaheuristic}                                                                                                        \\ \hline
		\multicolumn{1}{c|}{Name}         & \multicolumn{1}{c|}{Quantity} & \multicolumn{1}{c|}{Name}                                & \multicolumn{1}{c}{\begin{tabular}[c]{@{}c@{}}Quantity\\ (percentage)\end{tabular}} \\ \hline
		\multicolumn{1}{c|}{Ant}    & \multicolumn{1}{c|}{47}       & \multicolumn{1}{c|}{Ant Colony Optimization (ACO)}       & \multicolumn{1}{c}{24 (51.1\%)}                                                     \\ \hline
		\multicolumn{1}{c|}{Bat}    & \multicolumn{1}{c|}{46}       & \multicolumn{1}{c|}{Bat Algorithm (BA)}                  & \multicolumn{1}{c}{10 (21.7\%)}                                                     \\ \hline
		\multicolumn{1}{c|}{Bee}    & \multicolumn{1}{c|}{24}       & \multicolumn{1}{c|}{Artificial Bee Collony (ABC)}        & \multicolumn{1}{c}{14 (58.3\%)}                                                     \\ \hline
		\multicolumn{1}{c|}{DE}     & \multicolumn{1}{c|}{45}       & \multicolumn{1}{c|}{Differential evolution (DE)}         & \multicolumn{1}{c}{19 (42.2\%)}                                                     \\ \hline
		\multicolumn{1}{c|}{Evolutionary} & \multicolumn{1}{c|}{217}      & \multicolumn{1}{c|}{Genetic Algorithm (GA)}                   & \multicolumn{1}{c}{51 (23.5\%)}                                                     \\ \hline
		\multicolumn{1}{c|}{Heuristics}   & \multicolumn{1}{c|}{88}        & \multicolumn{1}{c|}{Heuristic Method (HM)}                                   & \multicolumn{1}{c}{5 (6\%)}                                                               \\ \hline
		\multicolumn{1}{c|}{GSL}          & \multicolumn{1}{c|}{3}        & \multicolumn{1}{c|}{Guided Local Search (GLS)}           & \multicolumn{1}{c}{2 (66.7\%)}                                                      \\ \hline
		\multicolumn{1}{c|}{HS}           & \multicolumn{1}{c|}{25}       & \multicolumn{1}{c|}{Harmony Search (HS)}                 & \multicolumn{1}{c}{13 (52.0\%)}                                                     \\ \hline
		\multicolumn{1}{c|}{ILS}          & \multicolumn{1}{c|}{28}       & \multicolumn{1}{c|}{Iterated Local Search (ILS)}         & \multicolumn{1}{c}{5 (17.9\%)}                                                      \\ \hline
		\multicolumn{1}{c|}{Neural}       & \multicolumn{1}{c|}{9}        & \multicolumn{1}{c|}{Artificial Neural Network (ANN)}                                   & \multicolumn{1}{c}{2 (22\%)}                                                               \\ \hline
		\multicolumn{1}{c|}{PSO}        & \multicolumn{1}{c|}{242}      & \multicolumn{1}{c|}{Particle Swarm Optimization (PSO)}   & \multicolumn{1}{c}{52 (21.5\%)}                                                     \\ \hline
		\multicolumn{1}{c|}{Simulated}    & \multicolumn{1}{c|}{58}       & \multicolumn{1}{c|}{Simulated Annealing (SA)}            & \multicolumn{1}{c}{29 (50.0\%)}                                                     \\ \hline
		\multicolumn{1}{c|}{Tabu}         & \multicolumn{1}{c|}{33}       & \multicolumn{1}{c|}{Tabu Search (TS)}                    & \multicolumn{1}{c}{12 (36.4\%)}                                                     \\ \hline
		\multicolumn{1}{c|}{VND}          & \multicolumn{1}{c|}{8}        & \multicolumn{1}{c|}{Variable Neighborhood Descent (VND)} & \multicolumn{1}{c}{2 (25\%)}                                                        \\ \hline
		\multicolumn{1}{c|}{VNS}          & \multicolumn{1}{c|}{18}       & \multicolumn{1}{c|}{Variable Neighborhood Search (VNS)}  & \multicolumn{1}{c}{4 (22.2\%)}                                                      \\ \hline
		\multicolumn{1}{c|}{Whale}        & \multicolumn{1}{c|}{11}       & \multicolumn{1}{c|}{Whale Optimization Algorithm (WOA)}  & \multicolumn{1}{c}{7 (63.6\%)}                                                      \\ \hline
		\multicolumn{1}{c|}{Wolf}        & \multicolumn{1}{c|}{46}       & \multicolumn{1}{c|}{Gray Wolf Optimizer (GWO)}  & \multicolumn{1}{c}{22 (47.8\%)}                                                      \\ \hline
	\end{tabular}
	\caption{Main metaheuristics of each family class and number of occurrences.}
	\label{tab:TableFrequenciaClassAndMainMetaheuristcs} 
\end{table}

\par
\begin{figure}[H]
	\centering
	\begin{tikzpicture}[scale=0.75]
		\begin{axis}[legend style={at={(1.1,0.5)},anchor=west},
			symbolic x coords={2004, 2005, 2006, 2007, 2008, 2009, 2010, 2011, 2012, 2013, 2014, 2015, 2016, 2017, 2018, 2019, 2020, 2021/22}, 
			xtick=data,width=12cm,height=7cm,
			ylabel={Number of Papers},ymin=0.0, ymax=55,
			x tick label style={rotate=90,anchor=east}]
			\addlegendentry{Ant}
			\addplot[mark=x,thick,red] coordinates {
				(2004,0) (2005,0) (2006,0) (2007,1) (2008,2) (2009,4) (2010,3) (2011,0) (2012,4) (2013,0) (2014,1) (2015,8) (2016,2) (2017,2) (2018,4) (2019,9) (2020,3) (2021/22,4)
			};
			\addlegendentry{BA}
			\addplot[mark=x,thick,black] coordinates {
				(2004,0) (2005,0) (2006,0) (2007,2) (2008,0) (2009,0) (2010,0) (2011,2) (2012,0) (2013,0) (2014,4) (2015,2) (2016,2) (2017,4) (2018,3) (2019,5) (2020,6) (2021/22,18)
			};
			\addlegendentry{Bee}
			\addplot[mark=x,thick,cyan] coordinates {
				(2004,0) (2005,0) (2006,0) (2007,0) (2008,0) (2009,0) (2010,0) (2011,0) (2012,0) (2013,0) (2014,0) (2015,3) (2016,2) (2017,0) (2018,2) (2019,4) (2020,5) (2021/22,8)
			};
			\addlegendentry{DE}
			\addplot[mark=+,thick,green] coordinates {
				(2004,0) (2005,0) (2006,0) (2007,0) (2008,1) (2009,0) (2010,0) (2011,0) (2012,0) (2013,0) (2014,1) (2015,5) (2016,5) (2017,3) (2018,6) (2019,9) (2020,5) (2021/22,10)
			};
			\addlegendentry{Evolutionary}
			\addplot[mark=+,thick,blue] coordinates {
				(2004,2) (2005,0) (2006,3) (2007,6) (2008,3) (2009,3) (2010,3) (2011,8) (2012,5) (2013,3) (2014,6) (2015,25) (2016,11) (2017,21) (2018,25) (2019,41) (2020,23) (2021/22,34)
			};
			\addlegendentry{Heuristic}
			\addplot[mark=+,thick,yellow] coordinates {
				(2004,0) (2005,0) (2006,2) (2007,1) (2008,2) (2009,1) (2010,0) (2011,1) (2012,2) (2013,2) (2014,1) (2015,12) (2016,7) (2017,9) (2018,6) (2019,15) (2020,11) (2021/22,16)
			};
			\addlegendentry{PSO}
			\addplot[mark=o,thick,orange] coordinates {
				(2004,0) (2005,0) (2006,2) (2007,1) (2008,1) (2009,4) (2010,0) (2011,2) (2012,2) (2013,3) (2014,5) (2015,12) (2016,17) (2017,21) (2018,32) (2019,54) (2020,38) (2021/22,48)
			};
			\addlegendentry{SA}
			\addplot[mark=o,thick,magenta] coordinates {
				(2004,0) (2005,0) (2006,2) (2007,4) (2008,3) (2009,1) (2010,1) (2011,1) (2012,2) (2013,2) (2014,1) (2015,10) (2016,4) (2017,8) (2018,3) (2019,10) (2020,6) (2021/22,0)
			};
			\addlegendentry{TB}
			\addplot[mark=o,thick,brown] coordinates {
				(2004,0) (2005,0) (2006,1) (2007,1) (2008,4) (2009,1) (2010,2) (2011,1) (2012,0) (2013,0) (2014,0) (2015,6) (2016,4) (2017,4) (2018,1) (2019,4) (2020,0) (2021/22,4)
			};
			\addlegendentry{Wolf}
			\addplot[mark=o,thick,gray] coordinates {
				(2004,0) (2005,0) (2006,0) (2007,0) (2008,0) (2009,0) (2010,0) (2011,0) (2012,0) (2013,0) (2014,0) (2015,4) (2016,0) (2017,2) (2018,4) (2019,9) (2020,6) (2021/22,21)
			};
			
			\node at (43.0,350.0) {\includegraphics[width=6cm]{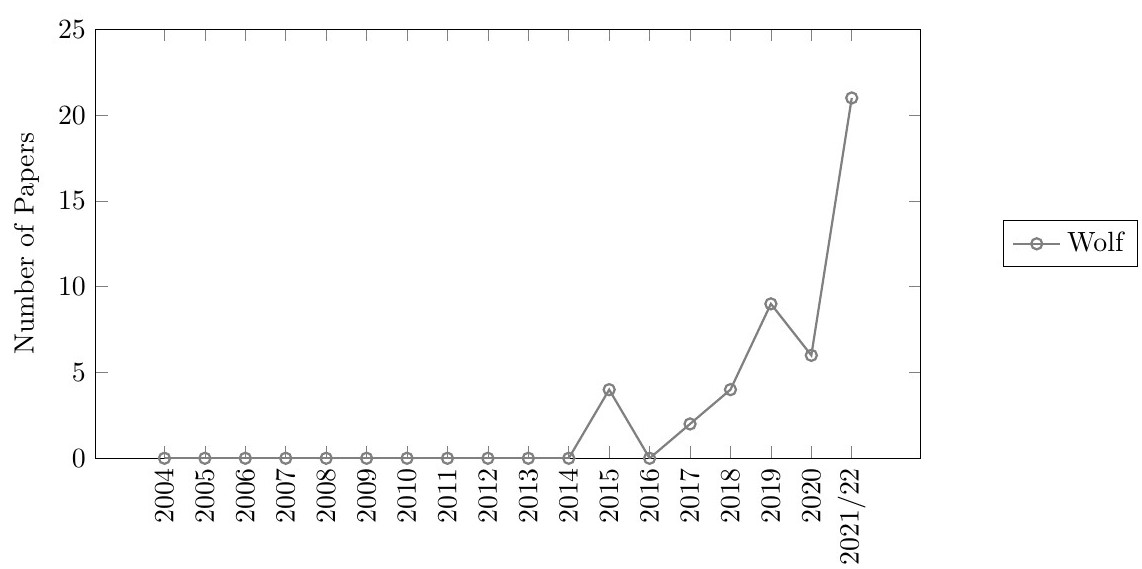}};
			,		\end{axis}
	\end{tikzpicture}
	\caption{Frequency of use of the family classes as a function of the year.}
	\label{fig:FrequenciaClassesMH}
\end{figure}
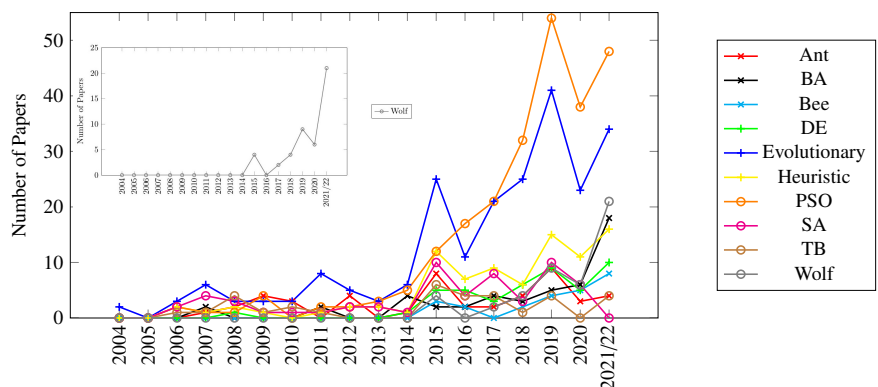

 Table \ref{tab:TableFrequenciaClassAndMainMetaheuristcs} evidence that the algorithms based on PSO and Evolutionary are intensively used to solve aerospace problems. Figure \ref{fig:FrequenciaClassesMH} illustrates the frequency of use of the family classes as a function of the year for the most relevant ones. Once again, it is evident that the use of the PSO and the Evolutionary based algorithms has increased rapidly. On the other hand, there is no evidence that the use of ILS, VNS, Whale, and VND families is increasing. Artificial intelligence-based solutions are poorly explored. There are references to their use only after 2015, but the citation of learning techniques seems to grow in the last three years of data collection. Notice, however, the general decrease in references for all family classes during the two years of the pandemic.

\par
\begin{figure}[H]
	\centering
	\begin{adjustbox}{width=\textwidth}
		\begin{tikzpicture}[scale=0.99,transform shape]
			\GraphInit[vstyle=Classic]
			\SetGraphUnit{3}
			\SetUpEdge[color = black,
			labelcolor = white,
			labeltext = red,
			labelstyle = {sloped,draw,text=blue}]
			\Vertex[Lpos=90]{Ant}
			\EA[Lpos=90](Ant){BA} \EA[Lpos=90](BA){Bee} \EA[Lpos=90](Bee){DE} \EA[Lpos=90](DE){Evolutionary}
			\SOEA(Evolutionary){GLS} \SO(GLS){Heuristic} \SO(Heuristic){HS} \SO(HS){ILS} \SOWE[Lpos=-90](ILS){Learning}
			\WE[Lpos=-90](Learning){Neural}
			\WE[Lpos=-90](Neural){PSO} \WE[Lpos=-90](PSO){SA} \WE[Lpos=-90](SA){TS} \NOWE[Lpos=180](TS){VND}
			\NO[Lpos=180](VND){VNS} \NO[Lpos=180](VNS){Whale} \NO[Lpos=180](Whale){Wolf}
			\tikzstyle{LabelStyle}=[fill=white]
			\tikzstyle{EdgeStyle}=[color=black]
			\Edge[lw = 1.85pt,label=$6$](PSO)(Evolutionary)
			\Edge[lw = 0.60pt,label=$1$](PSO)(SA)
			\Edge[lw = 0.60pt,label=$1$](PSO)(Bee)
			\Edge[lw = 0.60pt,label=$1$](PSO)(HS)
			\Edge[lw = 1.10pt,label=$3$](PSO)(DE)
			\Edge[lw = 0.60pt,label=$1$](PSO)(Neural)
			\Edge[lw = 3.45pt,label=$17$](PSO)(Heuristic)
			\Edge[lw = 0.60pt,label=$1$](PSO)(Learning)
			\Edge[lw = 1.35pt,label=$5$](PSO)(ILS)
			\Edge[lw = 0.60pt,label=$1$](Evolutionary)(SA)
			\Edge[lw = 0.60pt,label=$1$](Evolutionary)(HS)
			\Edge[lw = 0.60pt,label=$1$](Evolutionary)(Neural)
			\Edge[lw = 3.10pt,label=$14$](Evolutionary)(Heuristic)
			\Edge[lw = 2.35pt,label=$8$](Evolutionary)(Learning)
			\Edge[lw = 0.85pt,label=$2$](Evolutionary)(TS)
			\Edge[lw = 0.60pt,label=$1$](Evolutionary)(BA)
			\Edge[lw = 0.85pt,label=$2$](Evolutionary)(Wolf)
			\Edge[lw = 1.25pt,label=$4$](SA)(Heuristic)
			\Edge[lw = 0.85pt,label=$2$](SA)(VNS)
			\Edge[lw = 0.60pt,label=$1$](SA)(VND)
			\Edge[lw = 0.60pt,label=$1$](SA)(TS)
			\Edge[lw = 0.60pt,label=$1$](TS)(Ant)
			\Edge[lw = 1.35pt,label=$4$](TS)(Heuristic)
			\Edge[lw = 0.60pt,label=$1$](Ant)(DE)
			\Edge[lw = 0.60pt,label=$1$](Ant)(Heuristic)
			\Edge[lw = 0.60pt,label=$1$](Ant)(ILS)
			\Edge[lw = 0.60pt,label=$1$](Bee)(Wolf)
			\Edge[lw = 0.60pt,label=$1$](Bee)(Heuristic)
			\Edge[lw = 0.60pt,label=$1$](Bee)(ILS)
			\Edge[lw = 0.60pt,label=$1$](HS)(Heuristic)
			\Edge[lw = 0.60pt,label=$1$](BA)(Heuristic)
			\Edge[lw = 0.60pt,label=$1$](Wolf)(DE)
			\Edge[lw = 1.35pt,label=$4$](Wolf)(Heuristic)
			\Edge[lw = 0.60pt,label=$1$](Wolf)(ILS)
			\Edge[lw = 1.10pt,label=$3$](DE)(Whale)
			\Edge[lw = 1.85pt,label=$6$](DE)(Heuristic)
			\Edge[lw = 0.85pt,label=$2$](DE)(Learning)
			\Edge[lw = 0.60pt,label=$1$](Heuristic)(Learning)
			\Edge[lw = 0.85pt,label=$2$](Heuristic)(ILS)
			\Edge[lw = 0.60pt,label=$1$](Heuristic)(VNS)
			\Loop[dist=2.5cm,label=$1$](Heuristic)
			\Edge[lw = 0.60pt,label=$1$](Whale)(ILS)
			\tikzset{EdgeStyle/.style = {-,bend right}}
		\end{tikzpicture}
	\end{adjustbox}
	\caption{Binary hybrid metaheuristics.}\label{fig:AlgoritmosHibridos}
\end{figure}
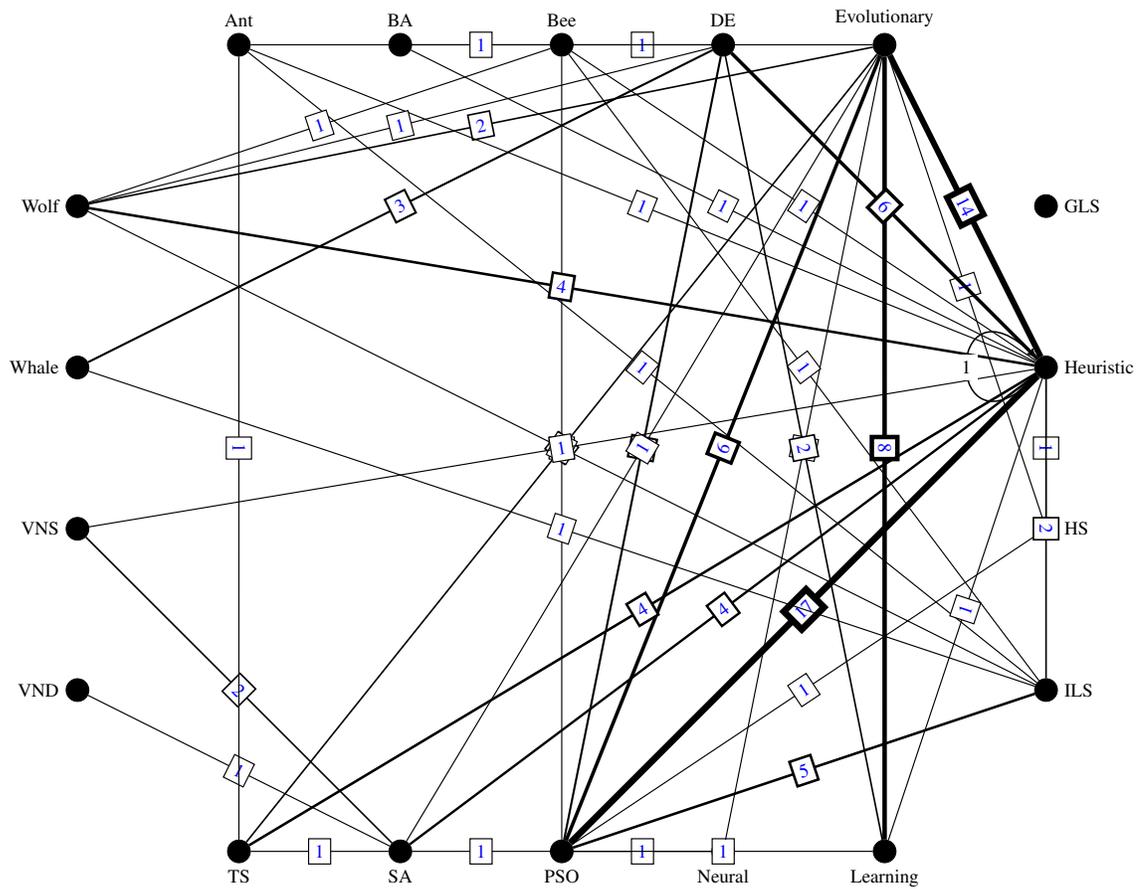

 Among the 377 algorithms, 31.56\% are hybrid metaheuristics. Hybrid metaheuristics can be classified into low-level or high-level hybridizations and are developed by combining two, three, or more algorithms. We classify the hybrids by the number of combined algorithms in the following. These combinations involve the composition of multiple metaheuristics and metaheuristics with heuristics or artificial intelligence-based methods. Hybrid metaheuristics combining two algorithms sum 111; seven hybrids are proposed combining three algorithms, and only one is generated from an association of four algorithms. Figure \ref{fig:AlgoritmosHibridos} presents the combinations of two algorithms found in the filtered articles and the number of times these combinations have occurred. Hybrid algorithms combining PSO-like metaheuristics and different heuristics are most frequently found in the articles, followed by associations between Evolutionary-like and Evolutionary metaheuristics, corresponding to 17 and 14 occurrences, respectively. In the following, we notice the associations between Evolutionary metaheuristics and learning techniques, with eight occurrences, DE + Heuristic, six occurrences, and PSO+ILS, five occurrences. Finally, observe the hybridization of PSO and evolutionary metaheuristics with six occurrences. It is interesting to highlight the four occurrences of hybridizations of the Wolf class, represented mainly by the Gray Wolf Optimizer and  Heuristics. Observing the timeline detail in Figure \ref{fig:FrequenciaClassesMH}, the Wolf family metaheuristics was first referenced in 2015 and shows an expressive increase of applications in aerospace problems. Only GLS was not used in binary hybridizations.

\par
\begin{figure}[H]
	\centering
	\begin{adjustbox}{width=\textwidth}
		\begin{tikzpicture}[scale=1.0,transform shape]
			\matrix[matrix of nodes, row sep=1cm, column sep=1cm] (m)
			{
				&         &  Evolutionary & GLS & Heuristic & HS & ILS & Learning & SA & PSO & TS & & \\
				&        &  \tikz\node[draw,circle,fill=black,scale=0.7]{6}; &  \tikz\node[draw,circle,fill=black,scale=0.7]{6}; & \tikz\node[draw,circle,fill=black,scale=0.7]{6}; & \tikz\node[draw,circle,fill=black,scale=0.7]{6}; & \tikz\node[draw,circle,fill=black,scale=0.7]{6}; & \tikz\node[draw,circle,fill=black,scale=0.7]{6}; & \tikz\node[draw,circle,fill=black,scale=0.7]{6}; & \tikz\node[draw,circle,fill=black,scale=0.7]{6}; & \tikz\node[draw,circle,fill=black,scale=0.7]{6}; & & \\
				Evolutionary & \tikz\node[draw,circle,fill=black,scale=0.7]{6}; &          &  &    &     &          &    &  &     &   & \tikz\node[draw,circle,fill=black,scale=0.7]{6}; & Evolutionary \\
				GLS           & \tikz\node[draw,circle,fill=black,scale=0.7]{6}; &        &    &    &     &          &    &     &  &   & \tikz\node[draw,circle,fill=black,scale=0.7]{6}; & GLS \\
				Heuristic    & \tikz\node[draw,circle,fill=black,scale=0.7]{6}; &          &  &    &     &          &    &     &    & & \tikz\node[draw,circle,fill=black,scale=0.7]{6}; & Heuristic \\
				HS           & \tikz\node[draw,circle,fill=black,scale=0.7]{6}; &        &    &    &     &          &    &    &   &   & \tikz\node[draw,circle,fill=black,scale=0.7]{6}; & HS \\
				ILS          &  \tikz\node[draw,circle,fill=black,scale=0.7]{6}; &        &    &    &     &          &    &    &    &  & \tikz\node[draw,circle,fill=black,scale=0.7]{6}; & ILS \\
				Learning     & \tikz\node[draw,circle,fill=black,scale=0.7]{6}; &         &   &    &     &          &    &   &     & & \tikz\node[draw,circle,fill=black,scale=0.7]{6}; & Learning \\
				SA           & \tikz\node[draw,circle,fill=black,scale=0.7]{6}; &      &      &    &     &          &    &   &   &    & \tikz\node[draw,circle,fill=black,scale=0.7]{6}; & SA  \\
				PSO        & \tikz\node[draw,circle,fill=black,scale=0.7]{6}; &       &     &    &     &          &    &    &   &   & \tikz\node[draw,circle,fill=black,scale=0.7]{6}; & PSO \\
				TS           & \tikz\node[draw,circle,fill=black,scale=0.7]{6}; &      &      &    &     &          &    &    &    &  & \tikz\node[draw,circle,fill=black,scale=0.7]{6}; & TS \\
				&            &  \tikz\node[draw,circle,fill=black,scale=0.7]{6}; & \tikz\node[draw,circle,fill=black,scale=0.7]{6}; &  \tikz\node[draw,circle,fill=black,scale=0.7]{6}; & \tikz\node[draw,circle,fill=black,scale=0.7]{6}; & \tikz\node[draw,circle,fill=black,scale=0.7]{6}; & \tikz\node[draw,circle,fill=black,scale=0.7]{6}; & \tikz\node[draw,circle,fill=black,scale=0.7]{6}; & \tikz\node[draw,circle,fill=black,scale=0.7]{6}; & \tikz\node[draw,circle,fill=black,scale=0.7]{6}; & & \\
				&       &  Evolutionary & GLS & Heuristic & HS & ILS & Learning & SA & PSO & TS & & \\
			};
			\draw (m-3-2) -- (m-2-10) -- (m-8-12);   
			\draw[orange] (m-10-2) -- (m-2-3) -- (m-5-12);   
			\draw[green] (m-9-2) -- (m-2-4) -- (m-11-12);   
			\draw[red] (m-5-2) -- (m-2-5) -- (m-8-12);   
			\draw[blue] (m-6-2) -- (m-2-5) -- (m-3-12) -- (m-12-10);   
		\end{tikzpicture}
	\end{adjustbox}
	\caption{Hybrid metaheuristics with 3 or 4 algorithm combinations.}
	\label{fig:AlgoritmosHibridos2}
\end{figure}
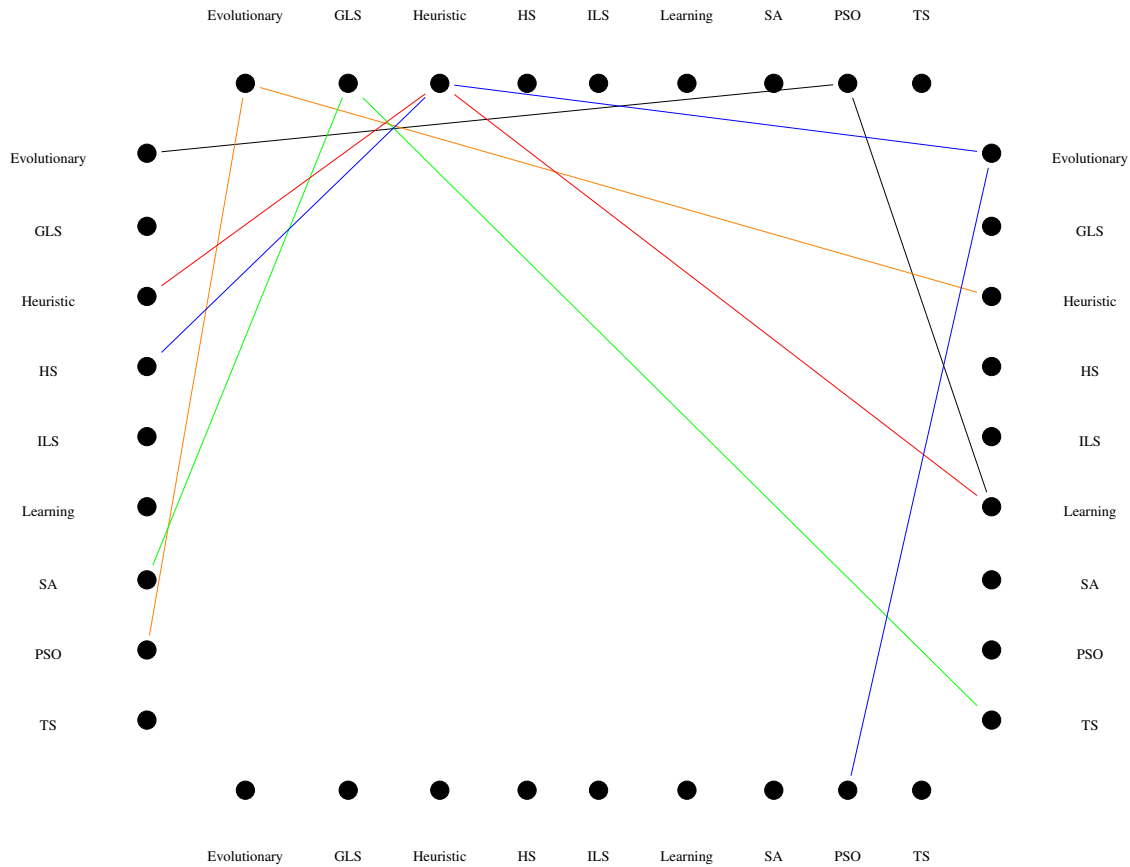

 Figure \ref{fig:AlgoritmosHibridos2} presents the hybridization of more than two heuristics and metaheuristics. Notice that GLS appears associated with SA and TS families. Again, hybridizations involving the PSO and Evolutionary families and associations of heuristics and learning techniques can be observed.

\par
\begin{figure}[H]
	\centering
	\begin{tikzpicture}[scale=0.7]
		\begin{axis}[legend style={at={(0.0,1.0)},anchor=west},
			symbolic x coords={2006, 2007, 2008, 2009, 2010, 2011, 2012, 2013, 2014, 2015, 2016, 2017, 2018, 2019, 2020, 2021,2022}, 
			xtick=data,width = 12cm, height = 6cm,
			ylabel={Number of Papers},ymin=-0.2, ymax=10,
			x tick label style={rotate=90,anchor=east}]
			\addlegendentry{New or Modified}
			\addplot[mark=*,thick,blue] coordinates {
				(2006,1) (2007,1) (2008,2) (2009,3) (2010,2) (2011,2) (2012,2) (2013,0) (2014,0) (2015,7) (2016,5) (2017,6) (2018,6) (2019,2) (2020,9) (2021,6) (2022,5)
			};
			\addlegendentry{Hybrid}
			\addplot[mark=+,thick,red] coordinates {
				(2006,0) (2007,2) (2008,0) (2009,0) (2010,2) (2011,1) (2012,4) (2013,1) (2014,3) (2015,2) (2016,4) (2017,2) (2018,4) (2019,6) (2020,6) (2021,8) (2022,1)
			};
		\end{axis}
	\end{tikzpicture}
	\caption{Focus of papers about the metaheuristics.}
	\label{fig:TendenciasPapers}
\end{figure}
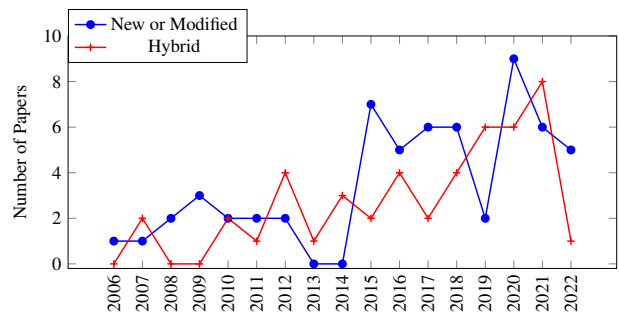

 Despite the small statistics, it is worth highlighting that the development of new algorithms and hybrid approaches for solving aerospace problems has been growing in recent years, Figure \ref{fig:TendenciasPapers}.

\par The articles can also be classified in terms of the number of objective functions approached:
\begin{enumerate}
	\item Mono-objective problems: 61.11\% of the articles focus on solving single-objective problems;
	\item Multi-objective problems: encompass 15.65\% of the papers; the majority of the multi-objective problems approached, 64.5\%, are bi-objective; three objectives are the focus of 25.8\% of the articles included in this class, and 9.7\%, just one problem, formulate the optimization problem with four objectives;
	\item Both: 21.20\% papers solve, or discuss, both single and multi-objective problems;
	\item  Not-specified: 2.04\%.
\end{enumerate}

\par Multi-objective optimization problems involving multiple conflicting objectives are present in scientific and engineering problems. These objectives must be optimized simultaneously to obtain a trade-off among the multiple objectives. Any metaheuristic can solve such problems allied to additional methods, such as the Weighted Sum Method \cite{2010Marler} or the épsilon-constraint methods \cite{2023Rahimi}. Moreover, many metaheuristics based on Pareto dominance have been developed and shown excellent performance in solving multi-objective problems \cite{2017Bao}. The majority of the problems solved in the filtered articles use regular metaheuristics, 93.53\%, and only 6.47\% use algorithms based on Pareto dominance, such as NSGA-II (16 occurrences) and MOGA (7 occurrences), that together represent 43.4\% of the multi-objective algorithms used in the selected articles.

\par Another important aspect to be explored when treating optimization problems is the type of optimization variable. The selected articles can be divided into four groups, according to the type of optimization variables of the treated problems, Table \ref{tab:TableVariables}.
\begin{table}[H]
	\centering
	\begin{adjustbox}{width=\textwidth}
		\begin{tabular}{ccc}
			\hline
			Problem type & Description & \% \\ \hline
			Continuous   & approaches problems with continuous decision variables only & 54.5\% \\
			Discrete     & uses metaheuristics to solve problems with discrete decision variables only & 37.4\% \\
			Mixed        & considers mixed optimization variables, both continuous and discrete & 5.6\%  \\
			Both         & solves more than one problem, but each problem with either integer or continuous variables only & 2.5\%  \\ \hline
		\end{tabular}
	\end{adjustbox}
	\caption{Main metaheuristics of each family class and number of occurrences.}
	\label{tab:TableVariables} 
\end{table}

\par
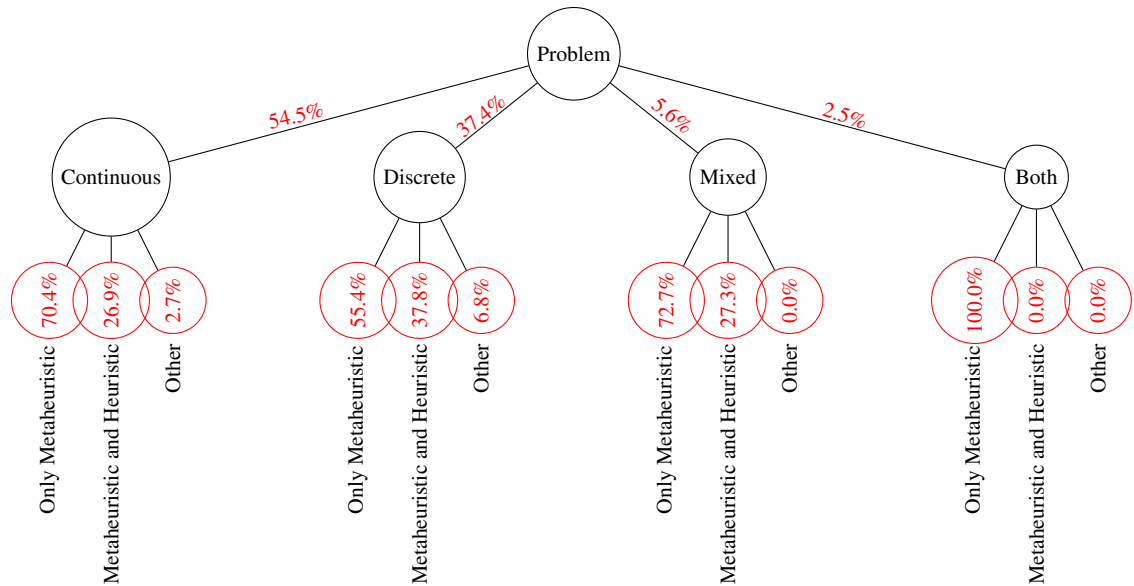
\begin{figure}[H]
	\centering
	\begin{tikzpicture}[level distance=2cm,
		level 1/.style={sibling distance=5.0cm},
		level 2/.style={sibling distance=1cm,level distance=2.0cm},
		scale=0.82,transform shape]
		\tikzstyle{every node}=[circle,draw]
		
		\node (Root) (A) {Problem}
		child {
			node {Continuous}
			child { node[red] {\begin{turn}{90}70.4\%\end{turn}} }
			child { node[red] {\begin{turn}{90}26.9\%\end{turn}} }
			child { node[red] {\begin{turn}{90}2.7\%\end{turn}} }
		}
		child {
			node {Discrete}
			child { node[red] {\begin{turn}{90}55.4\%\end{turn}} }
			child { node[red] {\begin{turn}{90}37.8\%\end{turn}} }
			child { node[red] {\begin{turn}{90}6.8\%\end{turn}} }
		}
		child {
			node {Mixed}
			child { node[red] {\begin{turn}{90}72.7\%\end{turn}} }
			child { node[red] {\begin{turn}{90}27.3\%\end{turn}} }
			child { node[red] {\begin{turn}{90}0.0\%\end{turn}} }
		}
		child {
			node {Both}
			child { node[red] {\begin{turn}{90}100.0\%\end{turn}} }
			child { node[red] {\begin{turn}{90}0.0\%\end{turn}} }
			child { node[red] {\begin{turn}{90}0.0\%\end{turn}} }
		};
		\node [draw=none,red, shift={(-4.5cm,-1.0cm)}] (A) {\begin{turn}{13.5}54.5\%\end{turn}};
		\node [draw=none,red, shift={(-1.5cm,-1.0cm)}] (A) {\begin{turn}{38}37.4\%\end{turn}};
		\node [draw=none,red, shift={(1.6cm,-1.0cm)}] (A) {\begin{turn}{-39}5.6\%\end{turn}};
		\node [draw=none,red, shift={(4.4cm,-1.0cm)}] (A) {\begin{turn}{-18.5}2.5\%\end{turn}};
		
		\node [draw=none,shift={(-8.5cm,-6.1cm)}] (A) {\begin{sideways}Only Metaheuristic\end{sideways}};
		\node [draw=none,shift={(-7.5cm,-6.65cm)}] (A) {\begin{sideways}Metaheuristic and Heuristic\end{sideways}};
		\node [draw=none,shift={(-6.5cm,-5.1cm)}] (A) {\begin{sideways}Other\end{sideways}};
		
		\node [draw=none,shift={(-3.5cm,-6.1cm)}] (A) {\begin{sideways}Only Metaheuristic\end{sideways}};
		\node [draw=none,shift={(-2.5cm,-6.65cm)}] (A) {\begin{sideways}Metaheuristic and Heuristic\end{sideways}};
		\node [draw=none,shift={(-1.5cm,-5.1cm)}] (A) {\begin{sideways}Other\end{sideways}};
		
		\node [draw=none,shift={(1.5cm,-6.1cm)}] (A) {\begin{sideways}Only Metaheuristic\end{sideways}};
		\node [draw=none,shift={(2.5cm,-6.65cm)}] (A) {\begin{sideways}Metaheuristic and Heuristic\end{sideways}};
		\node [draw=none,shift={(3.5cm,-5.1cm)}] (A) {\begin{sideways}Other\end{sideways}};
		
		\node [draw=none,shift={(6.5cm,-6.1cm)}] (A) {\begin{sideways}Only Metaheuristic\end{sideways}};
		\node [draw=none,shift={(7.5cm,-6.65cm)}] (A) {\begin{sideways}Metaheuristic and Heuristic\end{sideways}};
		\node [draw=none,shift={(8.5cm,-5.1cm)}] (A) {\begin{sideways}Other\end{sideways}};
	\end{tikzpicture}
	\caption{Class of hybrid metaheuristics with 3 and 4 algorithm combination.Distribution of papers by type of variable present in the problems versus the optimization procedure used to solve the model.}
	\label{fig:DiagramaDeArvore}
\end{figure}

 Still, focusing on continuous problems, 70.4\% are solved using only metaheuristics, 26.9\% use hybridized metaheuristics or just specialized heuristics, and 2.7\% use other mathematical or stochastic methods. For instance, in the last category, Salimi et al. \cite{169} use linear interpolation (LI) and spline interpolation (SI). A similar panorama is observed when considering articles solving problems modelled with other decision variable types, as illustrated in Figure \ref{fig:DiagramaDeArvore}.

\par The final statistics we approach is on the metaheuristics classification as populational or single-based solutions. Of the algorithms presented in papers, 54.71\% are population-based, 25.88\% of papers are single-based, and 19.41\% use mixed population/single-based metaheuristics.

\par In the next subsection, we presented an overview of the problems approached in the papers.

\subsection{Aerospace problems} \label{sec:Problems}

\par In this section, the focus is turned to the aerospace engineering problems solved in the analysed articles. Classifying the articles based on the problems solved, the following categories are defined: aerodynamic, benchmark problems, components, design problems, flight controller, flight parameter tests, fog computing, framework, human-like aerial manipulator, image enhancement, logistics, low earth orbit, structural analysis, surface detection, trajectory, timetabling, and others.

\par Table \ref{tab:ClassesDeProblemasVersusMetaheuristica} presents the relation between the used metaheuristics and the problem classes. We can highlight that the evolutionary and swarm classes are the most used to solve logistical problems. Notice that only in logistic problems all classes of metaheuristics were used. Furthermore, evolutionary and swarm were the classes of metaheuristics used in the most different classes of problems.
\begin{table}[H]
	\centering
	\caption{Frequency of problems versus metaheuristics presented in the papers analysed.}\label{tab:ClassesDeProblemasVersusMetaheuristica}
	\begin{adjustbox}{width=\textwidth}
		\begin{tabular}{ccccccccccccccccccl}
			\begin{turn}{90} Ant \end{turn} & \begin{turn}{90} Bat \end{turn} & \begin{turn}{90} Bee \end{turn} & \begin{turn}{90} DE \end{turn} & \begin{turn}{90} Evolutionary \end{turn} & \begin{turn}{90} GLS \end{turn} & \begin{turn}{90} Heuristic \end{turn} & \begin{turn}{90} HS \end{turn} & \begin{turn}{90} ILS \end{turn} & \begin{turn}{90} Learning \end{turn} & \begin{turn}{90} Neural \end{turn} & \begin{turn}{90} PSO \end{turn} & \begin{turn}{90} SA \end{turn} & \begin{turn}{90}\ TS \end{turn} & \begin{turn}{90} VND \end{turn} & \begin{turn}{90} VNS \end{turn} & \begin{turn}{90} Whale \end{turn} & \begin{turn}{90} Wolf \end{turn} &  \\
			1 & 1 & 0 & 1  &{\color{blue} 2}   & 0 & 0  & 0 & 0 & 0 & 1 &{\color{blue} 2}   & 0 & 1 & 0 & 0 & 0 & 0 & Aerodynamic \\
			5 & 5 & 1 & 10 &{\color{blue} 23}  & 0 & 17 & 0 & 0 & 1 & 0 &{\color{blue} 19}  & 2 & 5 & 0 & 4 & 1 & 3 & Benchmarks problems \\
			2 & 1 & 3 & 2  &{\color{blue} 16}  & 1 & 4  & 1 & 1 & 2 & 3 &{\color{blue} 21}  & 1 & 1 & 0 & 0 & 1 & 1 & Components \\
			0 & 0 & 0 & 0  &{\color{blue} 1}   & 0 & 0  & 0 & 0 & 0 & 0 &{\color{blue} 0}   & 2 & 1 & 0 & 0 & 0 & 0 & Design problems \\
			0 & 0 & 0 & 0  &{\color{blue} 2}   & 0 & 0  & 0 & 0 & 0 & 0 &{\color{blue} 2}   & 0 & 0 & 0 & 0 & 0 & 0 & Flight controller \\
			0 & 2 & 0 & 0  &{\color{blue} 4}   & 0 & 0  & 0 & 0 & 0 & 0 &{\color{blue} 1}   & 0 & 0 & 0 & 0 & 0 & 0 & Flight parameter tests\\
			0 & 1 & 0 & 0  &{\color{blue} 0}   & 0 & 0  & 0 & 7 & 0 & 0 &{\color{blue} 4}   & 0 & 0 & 0 & 0 & 1 & 1 & Fog computing\\
			0 & 0 & 0 & 0  &{\color{blue} 0}   & 0 & 1  & 1 & 0 & 0 & 0 &{\color{blue} 1}   & 1 & 0 & 0 & 0 & 0 & 0 & Framework\\
			0 & 0 & 0 & 1  &{\color{blue} 0}   & 0 & 0  & 0 & 0 & 1 & 0 &{\color{blue} 1}   & 0 & 0 & 0 & 0 & 0 & 0 & Human-like aerial manipulator\\
			0 & 0 & 0 & 0  &{\color{blue} 3}   & 0 & 0  & 0 & 0 & 0 & 0 &{\color{blue} 4}   & 0 & 0 & 0 & 0 & 0 & 0 & Image enhancement\\ 
			{\color{blue}24 }&{\color{blue} 13} &{\color{blue}11 } &{\color{blue}15 }  &{\color{blue} 113} & {\color{blue}2 } &{\color{blue}29 }  &{\color{blue} 7}  &{\color{blue}15 } &{\color{blue}15 } & {\color{blue}6 } &{\color{blue} 103} &{\color{blue} 30} &{\color{blue} 21} & {\color{blue} 6} &{\color{blue} 11} & {\color{blue} 2} &{\color{blue}10 } & Logistics\\
			0 & 0 & 0 & 0  &{\color{blue} 1}   & 0 & 0  & 0 & 0 & 0 & 0 &{\color{blue} 0}   & 0 & 0 & 0 & 0 & 0 & 0 & Low earth orbiting\\
			2 & 1 & 2 & 6  &{\color{blue} 13}  & 0 & 2  & 9 & 0 & 3 & 0 &{\color{blue} 7}   & 0 & 0 & 0 & 0 & 1 & 0 & Structural analysis\\
			0 & 1 & 0 & 0  &{\color{blue} 0}   & 0 & 1  & 0 & 0 & 0 & 0 &{\color{blue} 0}   & 1 & 0 & 0 & 0 & 0 & 0 & Surface detection\\
			2 & 4 & 2 & 3  &{\color{blue} 10}  & 0 & 9  & 0 & 3 & 1 & 0 &{\color{blue} 19}  & 4 & 3 & 0 & 2 & 1 & 1 & Trajectory\\
			0 & 0 & 0 & 0  &{\color{blue} 0}   & 0 & 1  & 0 & 0 & 0 & 0 &{\color{blue} 0}   & 0 & 0 & 0 & 0 & 0 & 0 & Timetabling\\
			6 & 1 & 0 & 2  &{\color{blue} 15}  & 0 &11  & 4 & 1 & 0 & 1 &{\color{blue} 14}  & 3 & 0 & 0 & 0 & 0 & 1 & Others\\
		\end{tabular}
	\end{adjustbox}
\end{table}

\par
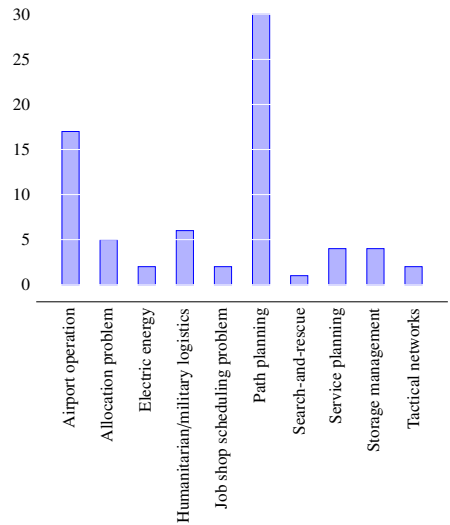
\begin{figure}[H]
	\centering
	\begin{tikzpicture}[scale=0.65]
		\begin{axis}[
			ybar,
			height=8cm,
			width=10cm,
			axis on top,
			y axis line style={opacity=0},
			axis x line*=bottom,
			tick align=inside,
			tick style={draw=none},
			xtick=data,
			ymajorgrids,
			major grid style={draw=white},
			xtick={1,2,3,4,5,6,7,8,9,10,11,12,13,14,15,16,17,18},
			xticklabels={Airport operation, Allocation problem, Electric energy, Humanitarian/military logistics, Job shop scheduling problem, Path planning, Search-and-rescue, Service planning, Storage management, Tactical networks},
			x tick label style={rotate=90,anchor=east},
			legend style={
				at={(0.0,-0.15)},
				anchor=east,
				legend columns=-1,
				/tikz/every even column/.append style={column sep=0.15cm}
			},
			]       
			\addlegendimage{empty legend}
			\addplot coordinates {
				(1, 17)
				(2, 5)
				(3, 2)
				(4, 6)
				(5, 2)
				(6, 30)
				(7, 1)
				(8, 4)
				(9, 4)
				(10, 2)
			};
		\end{axis} 
	\end{tikzpicture}
	\caption{Frequency of each problems classes of logistic presented in papers.}
	\label{fig:FrequenciaLogistica}
\end{figure}

 The logistics problems were subdivided into ten subgroups: airport operation, allocation problem, electric energy, humanitarian/military logistics, job shop scheduling problem, path planning, search-and-rescue, service planning, storage management, and tactical networks. The frequency distribution of the subclasses of the logistics problems is shown in Figure \ref{fig:FrequenciaLogistica}. In this case, the most prominent problems are airport operation and path planning.

\par The metaheuristics used to solve each subgroup are presented in Table \ref{tab:ClassesDeProblemasLogisticaVersusMetaheuristica}. All classes of metaheuristics defined in the previous section, Table \ref{tab:TableFrequenciaClassAndMainMetaheuristcs}, were explored to solve airport operation problems. The PSO class was used to solve all logistic problems, followed by the Evolutionary one, used in almost all subclasses but Job shopping scheduling and tactical networks. We highlight that airport operation is the most frequent class, with 40.48\% occurrences, followed by path planning with 28.57\%.
\begin{table}[H]
	\centering
	\caption{Frequency of logistics problems versus metaheuristics presented in the papers analysed.}\label{tab:ClassesDeProblemasLogisticaVersusMetaheuristica}
	\begin{adjustbox}{width=\textwidth}
		\begin{tabular}{ccccccccccccccccccl}
			\begin{turn}{90} Ant \end{turn} & \begin{turn}{90} Bat \end{turn} & \begin{turn}{90} Bee \end{turn} & \begin{turn}{90} DE \end{turn} & \begin{turn}{90} Evolutionary \end{turn} & \begin{turn}{90} GLS \end{turn} & \begin{turn}{90} Heuristic \end{turn} & \begin{turn}{90} HS \end{turn} & \begin{turn}{90} ILS \end{turn} & \begin{turn}{90} Learning \end{turn} & \begin{turn}{90} Neural \end{turn} & \begin{turn}{90} PSO \end{turn} & \begin{turn}{90} SA \end{turn} & \begin{turn}{90}\ TS \end{turn} & \begin{turn}{90} VND \end{turn} & \begin{turn}{90} VNS \end{turn} & \begin{turn}{90} Whale \end{turn} & \begin{turn}{90} Wolf \end{turn} &  \\
			{\color{blue} 8} &{\color{blue} 5}  &{\color{blue} 5}  &{\color{blue} 2}  &{\color{blue} 41}  &{\color{blue} 2}  &{\color{blue} 17} &{\color{blue} 4}  &{\color{blue} 12} &{\color{blue} 6}  &{\color{blue} 0}  &{\color{blue} 39}  &{\color{blue} 17} &{\color{blue} 1}  &{\color{blue} 6}  &{\color{blue} 10} &{\color{blue} 1}  &{\color{blue} 5}  & Airport operation \\
			0 & 1 & 0 & 2 &{\color{blue} 3}   & 0 & 2 & 0 & 0 & 0 & 1 &{\color{blue}  4}  & 0 & 0 & 0 & 0 & 0 & 0 & Allocation problem \\
			1 & 0 & 1 & 0 &{\color{blue} 18}  & 0 & 2 & 2 & 0 & 3 & 0 &{\color{blue} 16}  & 1 & 0 & 0 & 0 & 0 & 0 & Electric energy \\
			3 & 0 & 0 & 0 &{\color{blue} 5}   & 0 & 0 & 0 & 0 & 1 & 0 &{\color{blue} 1}   & 4 & 2 & 0 & 0 & 0 & 0 & Humanitarian/military logistics \\
			0 & 0 & 0 & 0 &{\color{blue} 0}   & 0 & 0 & 0 & 0 & 0 & 0 &{\color{blue} 2}   & 0 & 1 & 0 & 0 & 0 & 0 & Job shop scheduling problem\\
			{\color{blue} 8} &{\color{blue} 7}  &{\color{blue} 5}  &{\color{blue} 11} &{\color{blue} 31}  &{\color{blue} 0}  &{\color{blue} 5}  &{\color{blue} 1}  &{\color{blue} 2}  &{\color{blue} 2}  &{\color{blue} 5}  &{\color{blue} 35}  &{\color{blue} 4}  &{\color{blue} 2}  &{\color{blue} 0}  &{\color{blue} 0}  &{\color{blue} 1}  &{\color{blue} 5}  & Path planning\\
			0 & 0 & 0 & 0 &{\color{blue} 2}   & 0 & 0 & 0 & 0 & 0 & 0 &{\color{blue} 3}   & 0 & 0 & 0 & 0 & 0 & 0 & Search-and-rescue\\
			0 & 0 & 0 & 0 &{\color{blue} 3}   & 0 & 2 & 0 & 0 & 0 & 0 &{\color{blue} 1}   & 2 & 5 & 0 & 0 & 0 & 0 & Service planning\\
			1 & 0 & 0 & 0 &{\color{blue} 10}  & 0 & 0 & 0 & 1 & 3 & 0 &{\color{blue} 1}   & 2 & 0 & 0 & 1 & 0 & 0 & Storage management\\
			3 & 0 & 0 & 0 &{\color{blue} 0}   & 0 & 1 & 0 & 0 & 0 & 0 &{\color{blue} 1}   & 0 & 0 & 0 & 0 & 0 & 0 & Tactical networks\\
		\end{tabular}
	\end{adjustbox}
\end{table}

\subsection{Evaluation and validation of the filtering process}\label{sec:AIAAResults}

\par Analysing the selected articles allowed us to identify interesting information on the application of metaheuristics in the aerospace sector.
In the filtering process, articles from relevant scientific journals of the area were eliminated. In order to evaluate the impact of removing these articles from the analysis, additional evaluations were taken. For instance, articles from the AIAA repository were filtered because they do not have keywords section in the article body. These excluded articles were selected in this phase, and their content was evaluated. 

\par Eight articles compose this set \cite{2003Sousa,2011Ghisu,2013Kim,2017Min,2018Alam,2019Guo,2019Moura,2020Pant}. Five articles are classified as Performance Analysis \cite{2003Sousa,2017Min,2019Guo,2019Moura,2020Pant}, and three are case studies \cite{2011Ghisu,2013Kim,2018Alam}, a proportion compatible with the one found in Section \ref{sec:Papers}. Table \ref{tab:AIAASerch} shows the contribution of these articles in the context of Sections \ref{sec:Papers} and \ref{sec:Problems}. Regarding the metaheuristics' role in the papers, 50\% use well-established metaheuristics, 37.5\% propose and evaluate new metaheuristics, and 12.5\% (just one case) use a hybrid metaheuristic for a case study. Considering the low number of papers analysed in this section, the relation is close to the one presented in Figure \ref{fig:UsoDeMEtaheuristicas}.
\begin{table}[H]
	\centering
	\caption{Resume of analysis of articles from the AIAA repository that was filtered.}\label{tab:AIAASerch}
	\begin{adjustbox}{width=\textwidth}
		\begin{tabular}{ccccccc} \hline
			Reference        & Purpose     & Role            & Class of      & Number of   & Variables  & Class of \\
			&             &                 & Metaheuristic &  objectives &            & problems \\ \hline
			
			\cite{2003Sousa} & Performance & new             & Evolutionary  & 1           & discrete   & Design \\
			& Analysis    &                 &               &             &            & problem \\ \hline
			\cite{2011Ghisu} & Case        & well-stablished & TS            & 4           & continuous & aerodynamics \\
			& Study       &  metaheuristic  &               &             &            &              \\ \hline
			\cite{2013Kim}   & Case        & well-stablished & Evolutionary  & 3           & discrete   & Logistic - Airport \\
			& Study       &  metaheuristic  &               &             &            & operation  \\ \hline
			\cite{2017Min}   & Performance & new             & Learning      & 1           & continuous & Design \\
			& Analysis    &                 &               &             &            & problem \\ \hline
			\cite{2018Alam}  & Case        & Hybrid          & SA            & 1           & continuous & Logistic - Path \\
			& Study       &                 &               &             &            & planning        \\ \hline
			\cite{2019Guo}   & Performance & new             & Learning      & 1           & continuous & Components \\
			& Analysis    &                 &               &             &            &            \\ \hline
			\cite{2019Moura} & Performance & well-stablished & Evolutionary, & 1           & discrete   & Logistic - Service \\
			& Analysis    &  metaheuristic  & Learning, SA  &             &            &  planning \\ \hline
			\cite{2020Pant}  & Performance & well-stablished & Evolutionary  & various     & continuous & Design \\
			& Analysis    &  metaheuristic  &               &             &            &  problem \\ \hline
		\end{tabular}
	\end{adjustbox}
\end{table}

\par Regarding the number of objectives, the proportion is similar to that obtained in Section \ref{sec:Papers} for solving single-objective optimization problems,  62.5\%, and considering the low statistics is comparable for multi-objective problems,  25\% (two papers), and 12.5\% (one case) solve both single and multi-objective problems. The proportion of papers treating continuous variables is superior to the obtained with the selected papers after the filtering phase, which is expected because five articles deal with design problems, aerodynamics and components \cite{2003Sousa,2011Ghisu,2017Min,2019Guo,2020Pant}.

\par The eight articles considered in this section do not change the panorama presented in sections \ref{sec:Papers} and \ref{sec:Problems}.

\section{CONCLUSIONS}\label{sec:Conclusao}
\par This article presents a systematic study on using metaheuristics in the aerospace sector. We chose two sets of keywords for the searches, founding more than a thousand papers, and applied a set of filters to obtain a manageable number of papers for analysis. After filtering, 198 articles were selected for analysis.

\par The application of metaheuristics has grown in aerospace, mostly used in logistics. Several metaheuristics have been used, but the main ones are Evolutionary and Particle Swarm based metaheuristics. New or modified metaheuristics have been proposed, and several articles are devoted to evaluating their performance. One new metaheuristic, in particular, deserves mention, the Gray Wolf Optimizer, whose use has grown considerably since it was first used in 2015. Metaheuristics have been used for continuous and discrete optimization problems. However, a few papers solve mixed problems. 

\par Metaheuristics has been used to solve single and multi-objective problems, mostly bicriteria optimizations. Only in a few cases, Pareto dominance-based metaheuristics have been explored, leaving room for future research. A considerable number of papers, about a third, approach hybrid metaheuristics, mainly after 2014. However, data indicate a modest increase in using hybrid algorithms and artificial intelligence techniques in the area. There is plenty of room for additional research on the use and performance evaluation of hybrid algorithms in aerospace. Despite the small use of machine learning found in the papers consulted, using machine learning for enhancing metaheuristics seems to be promising research in the next years.

\par A considerable number of papers, about a third, approach hybrid metaheuristics, mainly after 2014. However, data indicate a modest increase in using hybrid algorithms and artificial intelligence techniques in the area. There is plenty of room for additional research on the use and performance evaluation of hybrid algorithms in aerospace. Despite the small use of machine learning found in the papers consulted, using machine learning for enhancing metaheuristics seems to be promising research in the next years.

\vspace*{-15pt}
\section*{ACKNOWLEDGEMENTS}
This work was supported in part by the Conselho Nacional de Desenvolvimento Científico e Tecnológico under Grant 307691/2020-9, and by the Coordenação de Aperfeiçoamento de Pessoal de Nível Superior - Brasil (CAPES) - Finance Code 001.

\section*{COMPETING INTERESTS}
The author(s) declare none.

\bibliographystyle{ieeetr}
\bibliography{references}

\end{document}